\RequirePackage{snapshot}
\documentclass[nofootinbib]{AIAArevtex4}

\usepackage{graphicx}
\usepackage{float}
\usepackage[caption=false]{subfig}
\usepackage[space]{grffile}
\usepackage{latexsym}
\usepackage{enumerate}
\usepackage{textcomp}
\usepackage{amsfonts,amsmath,amssymb}
\usepackage{bm}
\DeclareMathOperator*{\argmax}{\arg\!\max}
\usepackage{nomencl}%   nomenclature generation via makeindex
  \makenomenclature
\usepackage{url}
\usepackage{algorithm}
\usepackage{algpseudocode}
\usepackage{hyperref}
\usepackage{tabularx}
\hypersetup{colorlinks=false,pdfborder={0 0 0}}
\usepackage[utf8]{inputenc}
\usepackage{longtable}
\usepackage{multirow,booktabs,makecell}
\usepackage{color}
\newcommand{\BO}{GPBO}
\newcommand{\gp}{\mathcal{GP}(\fm,\fk)}
\newcommand{\fy}{y}
\newcommand{\fym}{y_m}
\newcommand{\K}[2]{K({#1},{#2})}
\newcommand{\Kx}{\K{X}{X}}
\newcommand{\Kxs}{\K{X_*}{X_*}}
\newcommand{\Kxxs}{\K{X}{X_*}}
\newcommand{\Kxsx}{\K{X_*}{X}}

\newcommand{\fv}{\bm{\mathrm{f}}}
\newcommand{\fs}{\mathcal{S}}
\newcommand{\fm}{m(\bm x)}
\newcommand{\fk}{k(\bm x,\bm x^\prime)}
\newcommand{\fa}{a(\bm x)}

\def\mr[#1]#2#3{\multirowcell{#2}[#1]{#3}}

\begin{document}
    \title{Adaptive Simulation-based Training of AI Decision-makers using Bayesian Optimization}

    \author{Brett Israelsen \footnote{Graduate Researcher, Computer Science, AIAA Student Member}}
    \author{Nisar Ahmed\footnote{Assistant Professor, Aerospace Engineering Sciences, AIAA Member}}
    \affiliation{University of Colorado, Boulder,~CO,~80309, USA}
    \author{Kenneth Center\footnote{Director, Orbit Logic Incorporated, AIAA Professional Member}}
    \author{Roderick Green\footnote{Sr. Software Engineer, Orbit Logic Incorporated, MD}}
    \affiliation{Orbit Logic Incorporated, Greenbelt,~MD,~20770, USA}
    \author{Winston Bennett Jr.\footnote{Division Technical Advisor -- Air Force Research Laboratory Warfighter Readiness Research Division}}
    \affiliation{Wright Patterson AFB,~OH,~45433, USA}

    \begin{abstract}
This work studies how an AI-controlled dog-fighting agent with tunable decision-making parameters can learn to optimize performance against an intelligent adversary, as measured by a stochastic objective function evaluated on simulated combat engagements. Gaussian process Bayesian optimization (GPBO) techniques are developed to automatically learn global Gaussian Process (GP) surrogate models, which provide statistical performance predictions in both explored and unexplored areas of the parameter space. This allows a learning engine to sample full-combat simulations at parameter values that are most likely to optimize performance and also provide highly informative data points for improving future predictions. However, standard GPBO methods do not provide a reliable surrogate model for the highly volatile objective functions found in aerial combat, and thus do not reliably identify global maxima. These issues are addressed by novel Repeat Sampling (RS) and Hybrid Repeat/Multi-point Sampling (HRMS) techniques. Simulation studies show that HRMS improves the accuracy of GP surrogate models, allowing AI decision-makers to more accurately predict performance and efficiently tune parameters. 
    \end{abstract}

    \maketitle
    % \printnomenclature

    \section{Introduction}
Rapid advancement in the capabilities of Artificial Intelligence (AI) has the potential to completely revolutionize the way that U.S. armed forces train for battlespace dominance. As AI becomes more sophisticated, there are many opportunities to insert AI into domain training environments. One use of AI is as the basis of agents that stand in for human vehicle/platform controllers in Live-Virtual-Constructive (LVC) simulations. Humans participating in training exercises in these environments need to be challenged at an appropriate level relative to their skills. Needed are agents that can assess the skill-level of human participants and adapt accordingly, serving as credible and adaptable adversaries that are indistinguishable from experienced humans. This work investigates how an AI agent with tunable parameters governing its overall behavior can be adapted to optimize an objective function for engagement outcomes.

Several challenges make it difficult to meet these objectives:    
    \begin{enumerate}
        \item Simulating an engagement can be costly. Beyond the financial expense of operating the simulation environment, contributions to the cost may also include the involvement of skilled personnel with limited availability, and the wall-clock duration of the simulation itself.\label{pnt:cost}
        \item The engagement metrics that need to be optimized cannot be described analytically, but can only be evaluated by running simulations. When performance evaluations are sampled, they are generally highly nonlinear functions of environmental parameters and decision maker states. Consequently, many traditional optimization methods are not applicable.\label{pnt:uglyobjective}
        \item The realistic nature of engagement simulations makes virtually all performance objective functions of interest extremely volatile and uncertain (e.g. due to combined random effects of weather, terrain, sensor noise, psycho-motor time delays, etc.).\label{pnt:volatile}        
    \end{enumerate}

Bayesian optimization with Gaussian Process surrogate models (abbreviated as \BO) is well-suited for directly addressing points \ref{pnt:cost} and \ref{pnt:uglyobjective}. In this formulation the Gaussian Process (GP) serves as a tractable surrogate model that approximates the true (non-closed form/intractable) objective function. This surrogate model estimates a nonparametric probability distribution over the objective function values at all location in the solution space. The Bayesian optimization algorithm uses this surrogate model to intelligently search the solution space for the optimum, based on a number of sampled function evaluations, i.e. using `explore/exploit' strategies to locate the minimum as quickly as possible while also using sparse function evaluations to build the surrogate model. Since the GP surrogate model is cheaper to evaluate than the true objective function, global nonlinear optimization methods can be used on the GP model to efficiently search the decision parameter space while also accounting for uncertainty in the underlying objective function. 

However, standard \BO{} methods are not well suited to address point \ref{pnt:volatile}. This work develops a novel approach for implementing \BO{}, called Hybrid Repeat/Multi-point Sampling (HRMS), to address these issues. In the setting of simulated one-on-one aerial dog-fighting engagements, \BO{} with HRMS is able to not only identify the optimum more reliably than standard \BO, but also yield a more accurate and consistent surrogate representation of the objective surface -- using no more total function evaluations that traditional \BO{} techniques. 

The remainder of this paper is outlined as follows. Section \ref{sec:background} discusses relevant prior work and provides a formal definition of the adaptive AI agent problem; relevant details regarding the application of \BO{} for aerial dog fighting performance optimization are also provided. Details of the proposed \BO{} `learning engine' framework used to train the decision-making AI are given in Section \ref{sec:methodology}, which also includes some discussion regarding practical implementation of \BO{}. Section \ref{sec:parallel_search} reviews existing sampling strategies for \BO{} and introduces our novel sampling strategy: HRMS.  Finally, Section \ref{sec:results} shows by simulated experiments that the proposed implementation of \BO{} with HRMS is useful for optimizing highly volatile performance metrics for the dog-fighting application. It is shown that HRMS can significantly outperform traditional \BO{} sampling techniques when dealing with highly volatile objective functions, and yields valuable insights about AI decision maker performance through the global GP surrogate model. 

    \section{Preliminaries and Problem Description}\label{sec:background}

\subsection{Problem Domain and Previous Work} \label{sec:ProbDomain}
    Logistical and fiscal constraints have led to a recent surge in interest around simulation-based methods for training warfighters. As such, efforts such as the Air Force's `Not-So-Grand Challenge' were developed with the specific goals of investigating solutions for current and future simulation training systems. As part of this effort, different autonomous decision making AI agents have been developed and evaluated based on their ability to effectively mimic human pilots in different situations~\cite{Doyle2014}. Although such AIs can effectively mimic human behavior to varying degrees of success in different circumstances, there still remains the question of whether they can adapt based on their adversaries' responses. The specific problem considered here is how a single autonomous agent with certain behavioral parameters can adapt to improve performance in response to other autonomous agents (human or AI) in a combat simulation featuring stochastic uncertainties and highly volatile outcomes. 

    %\subsubsection{Previous Work}
    The application focus is on simulations for air-to-air combat (dog-fighting) training, which have been studied extensively. For instance, McManus and Goodrich~\cite{McManus1990} discuss the integration of an AI-based tactical decision generator (TDG) system into two separate simulators to study and evaluate air combat environments. One of the simulation modes included an interface for human pilots to participate in training against the TDG. More recently, the `Not-So-Grand Challenge' produced multiple `human-like' AI systems to train against human pilots~\cite{Doyle2014}. Several different teams participated and were each rated on several metrics of performance and effectiveness against human pilots. State of the art techniques, such as  inverse reinforcement learning, hierarchical logic, and other proprietary approaches, were used to develop the AI pilots. One crucial factor for performance evaluation in this case is the degree to which the AI can mimic realistic human decision making. While it was found that some AI systems were more effective than others in this regard for specific scenarios (e.g. execution of evasive maneuvers, formation maintenance, target engagement, etc.), the question of how to efficiently `retune' and adapt any particular AI based on adversarial responses remained open. 
    
    This leads to the consideration of another critical component in the evaluation process for pilots (human or AI): development of suitable metrics that quantify performance during an engagement. For instance, Moore et al. discuss formal methods for measuring pilot dog-fighting performance and validated it in simulated combat scenarios~\cite{Moore1979}. These measures of pilot performance are meant to be less subjective than traditional ratings given by instructors/expert observers (as in the Not-So-Grand Challenge), as they are based on data recorded during an engagement. This touches on another motivation for developing AI pilots, i.e. automation of instructor functions and expert training resources, which are costly to provide and maintain. Ideally, if an autonomous AI can control a complex optionally manned vehicle such as a fighter plane, then it should also be able to evaluate and advise human pilots based on that same expertise (in much the same way human instructors are able to do this).  

    Performance metrics of aircraft engagement scenarios have evolved considerably since the inception of engagement debriefings. Many of these metrics are well accepted in the community. In contemporary development of newer metrics, expert evaluations are still utilized for validation. Kelly reviews and summarizes much of this work~\cite{Kelly1988}, and specifically mentions metrics that include variables such as: relative aircraft position, throttle and speedbrake manipulation, and overall engagement outcomes to name a few. Identification of meaningful metrics that operate on time-series and summary data from engagements is still an active field of research ~\cite{Kelly1988,Mulgund1998,Huynh1987,McManus1990,Mulgund2001,Paranjape2006}.  The work reported in this paper utilizes several metrics developed in the fighter pilot training literature as objective functions for automated learning and tuning of decision-making AI (although it is not exclusive to any particular set of metrics, or dog-fighting training applications, per se). 

    A large segment of work on optimization of aircraft engagement focuses on optimal teaming strategies. Mulgund et al. examined `large-scale' air combat tactics (formations, etc.) and were able to demonstrate promising results in that area~\cite{Mulgund1998,Mulgund2001}. Wu et al. addressed the problem of optimizing cooperative multiple target attack using genetic algorithms (GAs)~\cite{Wu2005}. Also applying GAs, Gonsalves and Burge investigated how mission plans could be optimized~\cite{Gonsalves2004}. While these are interesting and important application areas, the present work is focused on one-on-one engagements between autonomous adversaries (where one or both agents is an AI decision-maker), and providing an adversary that can be adaptive to the skills of the other pilot (human/AI). In other words, as opposed to modifying team tactics/strategies, this work focuses on optimally adapting the behaviors of individual agents. Another key difference from this work is that direct optimization strategies such as GAs are not well-suited to simulations that feature noisy environments and highly volatile outcomes. The approach proposed in this work overcomes this limitation by producing models of the objective function that account for these sources of uncertainty to produce more reliable performance optimization results. 

    More recently, Ernest et. al designed an AI system that can function in real time during aerial combat ~\cite{Ernest2016}. Their work utilized genetic optimization of a group of Fuzzy Inference Systems (FISs) that together produce control actions given input data from the simulated environment. This system is able to produce `fine-grained' control actions of a single jet or groups of fighters. Also, due to the specification of the FISs, the control outputs are interpretable to humans. As in the present paper, the work of~\cite{Ernest2016} seeks an optimal solution to a combat scenario by modifying fighter behaviors. However, a key difference is that their approach requires designing the entire AI system from the top down, whereas the approach advocated in this paper operates on top of \textit{any} AI with tunable behavior parameters. Furthermore, the approach presented here tries to minimize the number of simulations needed to optimally adapt the AI's behavior to maximize performance, while building an easily interpretable global statistical model of the AI parameter-to-performance metric mapping at the same time. %This model can be used later by the AI to analyze/predict performance or initiate further adaptations. 

    Autonomous AI agent technology has also been extensively developed for the video game industry, where non-player characters (NPCs) interact with human players. Still, the majority of agent AIs used in the gaming industry are highly scripted and non-adaptive ~\cite{Yannakakis2012}. Cole et al. used GAs to tune game agents for first person shooter games~\cite{Cole2004a}. Liaw et al. used GAs to evolve game agents that work as a team~\cite{Liaw2013} . Othman et al. discuss using simulations to evolve an AI agent for tactical purposes~\cite{Othman2012a}. Similar to the literature on optimization in aircraft engagement, these methods do not account for the cost of evaluating objective functions for optimization (e.g. number of human experiments, or more generally financial/computational expense). They also do not attempt to construct a model of the true objective function (with uncertainty quantification) to enable better robustness and adaptation to the uncertainties involved with predicting and evaluating the outcome of real-time competitions between human/AI decision agents.

%%Nisar: Doesn't make sense to have this here any more: moving this to top of Section III
    %%Bayesian optimization has emerged as a critical tool for tuning hyperparameters in various areas of machine learning~\cite{Bergstra2011,Snoek2012,Mahendran2012}. It has also been applied in several other fields such as modeling of user preferences, and reinforcement learning~\cite{Brochu2010}. It is well suited for optimizing objective functions that are unknown and expensive to evaluate, and is often able to do this with the fewest function evaluations as compared to other competing methods~\cite{Jones1998}. These properties make it an ideal candidate for optimizing the behavioral parameters of an AI pilot. As mentioned earlier (and as will be demonstrated later), standard application of \BO{} is not able to perform well on objective functions that are highly volatile, or when an accurate model of the true objective function is desired. To this end we propose and demonstrate a novel sampling approach called Hybrid Repeat/Multi-point Sampling (HRMS) that not only yields more repeatable optimization results, but has more accurate representation of the true objective surface when the optimization is completed. 
    %\nomenclature{HRMS}{Hybrid Repeat/Multi-point Sampling}

    \subsection{Formal Problem Definition}\label{sec:problem_definition}
For simplicity and ease of developing the underlying theory, this work focuses on the problem of adjusting the parameters of a single AI decision maker competing against another AI decision maker in an aerial combat simulation. 
In this case, the `blue' agent represents the AI whose parameters are to be optimized, while the `red' agent represents an adversary whose decision making parameters remain fixed (but are not necessarily known to blue). The red agent in this case can also be viewed as a proxy for a human pilot in a training scenario. 
The decision-making parameter vectors are given by $\bm{x}_{r}$ and $\bm{x}_{b}$ for red and blue, respectively. 
For the aerial combat domain considered here, $\bm{x}_r$ and $\bm{x}_b$ are assumed to consist of variables such as those shown in Table \ref{tab:agent_params}. These parameters are taken from a proprietary aerial combat simulation AI developed by OrbitLogic, Inc. and define different threshold settings for triggering autonomous pilot behaviors defined by a state machine.  
These variables define the space over which optimization will occur in this paper, although other AI and decision making parameters could also be used. 

    % \usepackage{booktabs}
    \begin{table}[!htbp]
    \centering
    \caption{Agent Behavior Parameter elements ${\bf x}_b(j)$ }
    \label{tab:agent_params}
    \resizebox{\textwidth}{!}{%
    \begin{tabularx}{1.15\linewidth}{lllX}
    \toprule
    $j$ & \textbf{Abbreviation} & \textbf{Parameter Name} & \textbf{Description} \\ \midrule
    1 & attackrange & AttackRange & Distance at which attacker begins active pursuit of detected adversary \\
    2 & endspeed & EndSpeed & Speed to use while on patrol or acquiring ground targets \\
    3 & fixedvertical & FixedVertical & Altitude to use while on patrol or acquiring ground targets \\
    4 & intspeed & InterceptSpeed & Speed to use while in attack posture approaching adversary \\
    5 & launch & LaunchDelay & Time to wait to fire weapon after lock has been achieved \\
    6 & maxaz & MaxAximuth & Max azimuth to allow adversary when in pursuit \\
    7 & maxel & MaxElevation & Max elevation to allow adversary when in pursuit \\
    8 & minagl & MinAgl & Minimum above ground level (AGL) altitude when engaged in combat \\
    9 & select & SelectDelay & Delay in selecting weapon to employ \\
    10 & track & TrackDelay & Delay in responding to sensor indication of adversary track \\ \bottomrule
    \end{tabularx}
    }
    \end{table}

The goal is to minimize some desired engagement performance objective function $y_m(\bm{x}_{r},\bm{x}_{b})$ with respect to $\bm{x}_{b}$ while $\bm{x}_{r}$ is held constant. Here, $y_m(\cdot)$ must be evaluated using a high-fidelity combat simulation. The subscript $m$ denotes that there may be multiple objective functions for a single objective simulation environment $y$. Table \ref{tab:objective_metrics} lists some of the possible objective metrics for the simulation, which were compiled from the aforementioned literature on air combat training evaluations. Each of the metrics describes a post hoc evaluation of blue's performance against red, based on telemetry data, control input data, and other relevant dynamic variables recorded during each engagement. Although each of the parameters and the outcome metrics are well-defined, each simulation is expensive. From Table \ref{tab:agent_params} we also see that the optimization can take place in a fairly high-dimensional space. 

    \nomenclature{$\bm{x}_b,\bm{x}_r$}{Behavioral parameter vectors for red and blue agents}
    \nomenclature{$y_m(\bm{x}_{r},\bm{x}_{b})$}{Simulation objective function}

    % \usepackage{booktabs}
    % \usepackage{graphicx}
    \begin{table}[!htbp]
    \centering
    \caption{Simulation Objective Metrics $y_m(\cdot)$}
    \label{tab:objective_metrics}
    \resizebox{\textwidth}{!}{%
    \begin{tabularx}{1.15\linewidth}{lllX}
    \toprule
    $m$ & \textbf{Abbreviation} & \textbf{Metric Name} & \textbf{Description} \\ \midrule
    1 & survived & Survived & Boolean representing whether the agent survived or not \\
    2 & TST & TotalSimulatedTime & Duration of the simulation \\
    3 & numKills & NumberOfKills & Number of adversary kills achieved \\
    4 & MOC & MissionObjectivesCount & Number of ground targets acquired \\
    5 & cTTK & CumulativeTimeToKill & Cumulative time between weapons lock and target elimination \\
    6 & cTOO & CumulativeTimeOnOffnese & Cumulative time in pursuit \\
    7 & cTOD & CumulativeTimeOnDefense & Cumulative being pursued \\
    8 & WU & WeaponsUsed & Number of weapons employed during the simulation \\
    9 & cTAL & CumulativeTimeToAchieveLock & Cumulative time between detection of adversary and weapons lock \\
    10 & EMM & EnergyManagementMetric~\cite{Kelly1979a} & Integrated total energy delta relative to adversary in the course of the simulation \\
    11 & SAM & StickAggressivenessMetric~\cite{Kelly1979a} & Measure of aggressiveness applied to control of the aircraft in the course of the simulation \\ \bottomrule
    \end{tabularx}%
    }
    \end{table}

\begin{figure}[!htbp]
          \centering
          \includegraphics[width=0.85\textwidth]{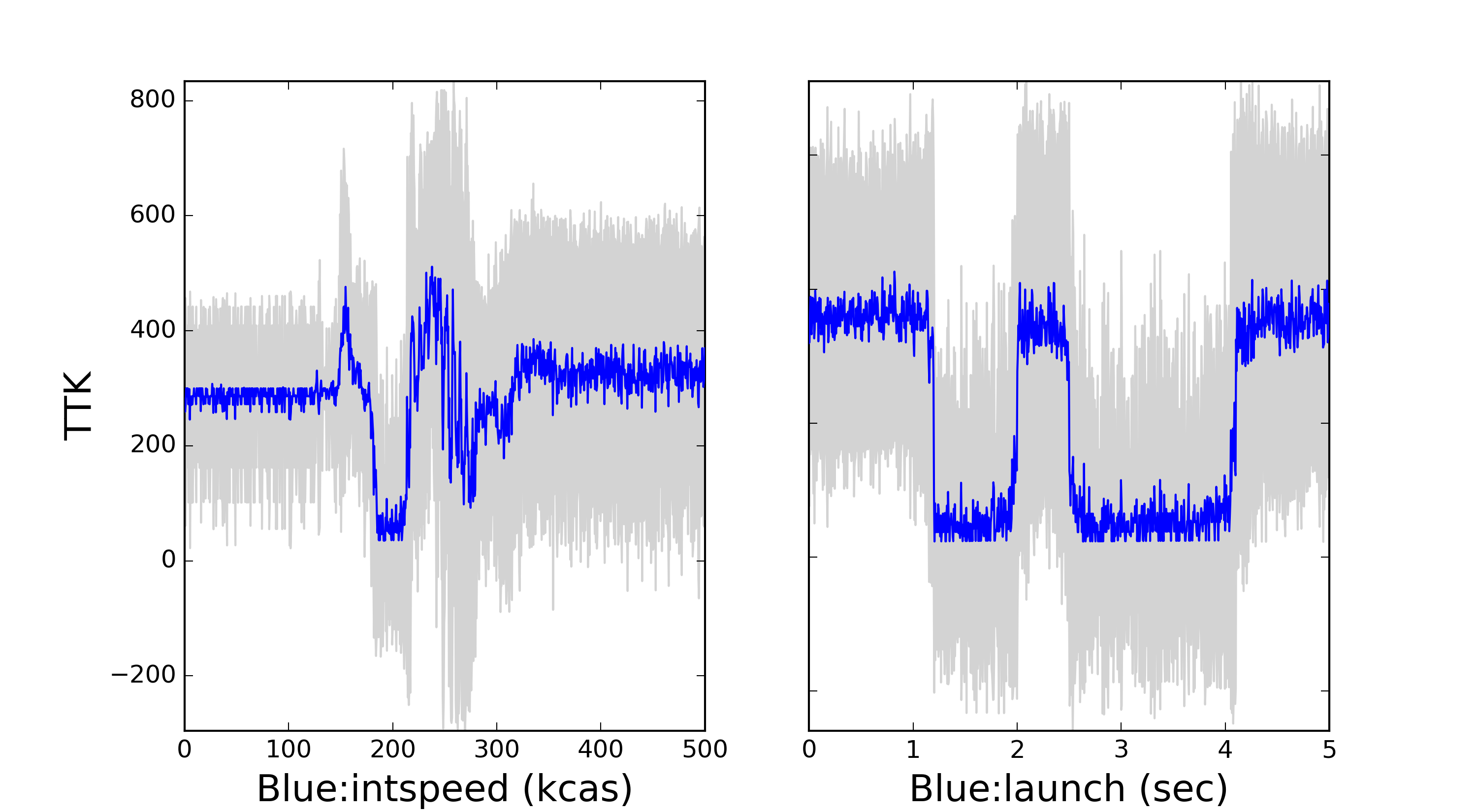}
          \caption{Simulation examples of the \emph{TTK} objective function vs. two different $\bm{x}_b$ variables \emph{Blue:intspeed}, \emph{Blue:launch}. The mean and 2$\sigma$ uncertainty bounds are shown as dark blue line and the shaded area, respectively. Data was obtained by sampling the objective function 20 times at 1000 locations across the horizontal axis.}
          \label{fig:objective_examples}
    \end{figure}

    Figure \ref{fig:objective_examples} shows some examples of $y_m(\cdot)$ for typical simulation runs of the one-on-one engagement scenario defined later in Section III. These figures were produced by holding all ${\bm{x}_r}$ parameters constant, as well as all $\bm{x}_b$ parameters except the one listed. Data was obtained by sampling the simulated engagement 1000 times along the horizontal axis and repeating that process 20 times. Clearly, these performance functions are highly noisy and volatile due to stochastic uncertainties. These uncertainties are due to environmental conditions such as, wind speed, terrain, and weather, as well as other sources such as error from sensors, psycho-motor time delays, and initial placement of the fighters. Furthermore, it is clear that $y_m(\cdot)$ will generally not be a convex function of $\bm{x}_b$. Consequently, a specialized class of optimization is needed that can: (i) handle inherently noisy $y_m(\cdot)$ data; (ii) identify the best local minima with respect to $\bm{x}_b$; and (iii) reduce the number of evaluations needed to find the (expected) minimum of $y_m(\cdot)$. Bayesian optimization will now be introduced as a tool that can address these inherent challenges. 
    
    \subsection{Bayesian Optimization}\label{sec:BayesOpt}
The goal of optimization is to minimize some objective function $y:\mathcal{X} \rightarrow \mathbb{R}$, where $\mathcal{X} \in \mathbb{R}^d$ is the search or solution space, and the element $\bm{x}^* \in \mathcal{X}$ is the minimizer, such that $y(\bm{x}^*) \leq y(\bm{x}), \  \forall \bm{x} \in \mathcal{X}$. Typically the solution space is bounded for global optimization, where $\bm{x}(i) \in [x(i)_l, x(i)_u]$ for lower bound $x(i)_l$ and upper bound $x(i)_u$ for element $i$ of $\bm{x}$. In classical optimization theory, one assumes the mathematical description of $y$ as a function of $\bm{x}$ is exactly known, e.g. $\fy = x^{2} + 3x^{3}$ for scalar $\bm{x}=x$. In this case, it is fairly inexpensive/fast to evaluate $y$ and apply standard optimization techniques. However, when the mapping from $\bm{x}$ to $\fy$ is not known explicitly, then optimization typically requires the evaluation of a `black box' function. In many applications, such as the one considered here (i.e. where $\fy$ is a high-fidelity simulator, the output of which is final engagement metrics), the black box evaluations of $\fy$ can be expensive, slow, and produce noisy results for the same input values $\bm{x}$.
% \nomenclature{$y$}{objective function}
% \nomenclature{$\mathcal{X}$}{optimization search space} \nomenclature{$\mathbb{R}^d$}{d-dimensional Euclidean space}
% \nomenclature{$d$}{dimension of the search space}
% \nomenclature{$\bm{x}^*$}{optimum in $\mathcal{X}$}

The goal of Bayesian optimization is to find the minimizer of some unknown noisy objective function that is costly to evaluate, while also learning about it at the same time. It is so named because Bayesian inference, a key concept from  probability theory, is applied during the optimization process. That is, the process uses an initial prior belief $p(y)$ over potential $\fy$ functions, which can be updated by subsequent observations or evidence $E$ consisting of sample evaluations of $\fy$ for different values of $\bm{x}$. Mathematically, this leads to an application of Bayes' rule: $p(y|E) \propto p(E|y)p(y)$ (see appendix A.1 of~\cite{Rasmussen2006} for more detail). The quantity $p(E|y)$ is also known as the observation likelihood, while $p(y|E)$ is the posterior, i.e. the updated probability of $\fy$ after $E$ is observed. Hence, given some initial prior belief about $y$ and evidence $E$ about the actual shape of $\fy$, an updated posterior belief about $\fy$ can be formed. As long as both the prior and likelihood are consistent with the true nature of $\fy$, then the law of large numbers ensures that the posterior $p(y|E)$ converges with high probability to the true $\fy$ in the limit of infinite observations $E$ covering $\mathcal{X}$. 

% \nomenclature{$p(y|E)$}{probability of $y$ given evidence $E$}

Bayesian optimization uses black box point evaluations of $y$ to find $\bm{x}^*$. This is accomplished by maintaining beliefs about how $\fy$ behaves over all $\bm{x}$ in the form of a `surrogate model' $\fs$, which statistically approximates $\fy$ and is easier to evaluate (again $\fy$ might be an expensive high fidelity simulation). During optimization, $\fs$ is used to determine where the next point sample evaluation of $\fy$ should occur, in order to update beliefs over $\fy$ and thus simultaneously improve $\fs$ while finding the (expected) minimum of $\fy$ as quickly as possible. The key idea is that, as more observations are sampled at different $\bm{x}$ locations, the $\bm{x}$ samples themselves eventually converge to the expected minimizer $\bm{x}^*$ of $\fy$. Since $\fs$ contains statistical information about the level of uncertainty in $\fy$ (i.e. related to $p(y|E)$, the posterior belief), Bayesian optimization effectively leverages probabilistic `explore-exploit' behavior to learn an approximate model of $\fy$ while also minimizing it. The main components of the Bayesian optimization process are:
\begin{enumerate}
    \item a surrogate model $\fs$, which encodes statistical beliefs about $\fy$ in light of previous observations and a prior belief;
    \item an acquisition function $\fa$, which is used to intelligently guide the search for $\bm{x}^*$ via $\fs$.
\end{enumerate}
\label{bayesopt}
    \subsubsection{The Surrogate Model: }\label{sec:gaussian-processes}
The surrogate model $\fs$ must fulfill some basic properties to be useful in Bayesian optimization. First, it must approximate the function in areas where it has not yet been evaluated. Second, when it is evaluated it must yield both a predicted value and a corresponding uncertainty to quantify the possibility that the optimum is located at some location $\bm{x}$. Several different kinds of surrogate models can be used, e.g. Gaussian Processes (GPs)~\cite{Rasmussen2006}, Random Forests~\cite{Hutter2011}, and Neural Networks ~\cite{Snoek2015}. Different types of surrogate models are also available for categorical/integer inputs and for problems where $\bm{x}$ is very high-dimensional \cite{Gelbart2015,Snoek2015}. GPs are by far the most common model used as a surrogate model in Bayesian optimization, and are the focus of the work here. They are especially useful for continuous optimization problems of relatively low dimensionality ($d\approx10$)~\cite{eggensperger2013}. Henceforth, the acronym \BO{} refers to Bayesian optimization using a GP surrogate model $\fs$. 

%\paragraph{Gaussian Process Definition: }\label{definition-of-a-gp}
A GP can be used to describe a distribution over functions; it is more formally defined as a collection of random variables, any finite number of which have a joint Gaussian distribution (see equation~\ref{eq:gp_basic})~\cite{Bishop2006,Rasmussen2006}. The process is completely specified by its mean function $\fm$ (equation~\ref{eq:gp_mean}), and its covariance function $\fk$ (equation~\ref{eq:gp_cov}), 
    \begin{align}
        f(\bm x) &\sim \gp \label{eq:gp_basic} \\
        \fm &= \mathbb{E}[f(\bm{x})] \label{eq:gp_mean} \\
        \fk &= \mathbb{E}[(f(\bm{x})-m(\bm{x}))(f(\bm{x}^\prime)-m(\bm{x}^\prime))] \label{eq:gp_cov}
    \end{align}
In theory $\fm$ could be any function; as is common practice, this work assumes $m$ is zero for ease of exposition. The covariance function, also called a kernel, is a function that maps $\bm{x}$ and $\bm{x}^\prime$ to a scalar, or $k:(\bm{x},\bm{x}^\prime) \rightarrow \mathbb{R}$. These functions must be specified a priori, and would ideally be based on some prior knowledge of the properties of $\fy$, such as its smoothness; the kernel used in our research is presented in section \ref{sec:MLE_MAP_CV}.

A kernel must be positive semi-definite (PSD) to be valid. A kernel is said to be PSD if it produces a Gram matrix $K$, with individual elements $[K_{i,j}]$ given by $k(\bm{x}_{i},\bm{x}_{j})$ that is PSD given a set of training data $X = \{\bm{x}_{1},\ldots,\bm{x}_n\}$. Kernels are further discussed in Section \ref{sec:MLE_MAP_CV}. %\nisarcomm{missing !!!!}
$\Kx$ is the Gram matrix defined by kernel $k$, 
    \nomenclature{$\Theta$}{hyperparameters of GP kernel}
    \begin{align}
        \Kx{} &= \begin{bmatrix}
                k({\bm x}_1,{\bm x}_1) & k({\bm x}_1,{\bm x}_2) & \cdots & k({\bm x}_1,{\bm x}_n)  \\
								k({\bm x}_2,{\bm x}_1) & k({\bm x}_2,{\bm x}_2) & \cdots & k({\bm x}_2,{\bm x}_n)  \\
								\vdots		 &  \vdots    &   \vdots    \\ 
                k({\bm x}_n,{\bm x}_1) & k({\bm x}_n,{\bm x}_2) & \cdots & k({\bm x}_n,{\bm x}_n)
					   \end{bmatrix}
    \end{align}
Given $n$ training observations, the individual elements of the covariance matrix $\Kx \in \mathbb{R}^{n\times n}$ are the covariances $k({\bm x}_i,{\bm x}_j)$ between ${\bm x}_i$ and ${\bm x}_j$ for all pairs of training data. 
For Gaussian observation likelihoods, the joint distribution of $n$ training outputs $\fv(X)\in\mathbb{R}^{n\times 1}$ and $p$ test outputs $\fv_{*}(X_*)\in\mathbb{R}^{p\times 1}$ for $p$ test inputs $\{X_*=\bm{x}_{*1},\ldots,\bm{x}_{*p}\}$  can be written as 
	\begin{align}
        \begin{bmatrix}
            \fv \\
            \fv_* \end{bmatrix} &\sim  \mathcal{N}\left(\textbf{0},\begin{bmatrix}
                \Kx & \Kxxs \\
                \Kxsx & \Kxs \end{bmatrix}\right) \label{eq:fX}, \\
						\Kxsx & = \begin{bmatrix} 
						k(x_{*1},x_1) & k(x_{*1},x_2) & \cdots & k(x_{*1},x_n) \\
						k(x_{*2},x_1) & k(x_{*2},x_2) & \cdots & k(x_{*2},x_n) \\
						\vdots & \vdots & \ddots & \vdots \\
						k(x_{*p},x_1) & k(x_{*p},x_2) & \cdots & k(x_{*p},x_n) \end{bmatrix}
    \end{align}
Thus, given observed training values $X$ and $\fv$, predictions of $\fv_*$ can be made at new input locations $X_*$. 
For inference, the conditional values of the test outputs are of interest:
    \begin{align}
        \fv_{*}|X_{*},X,\fv &\sim \mathcal{N}(\mu(X_*),\sigma^2(X_*)) \label{eq:predict2} \\
        \mu(X_*) &= \Kxsx \Kx^{- 1}\fv \label{eq:mu}\\
        \sigma^2(X_*) &= \Kxs - \Kxsx \Kx^{- 1} \Kxxs \label{eq:sigma}
    \end{align}
    \nomenclature{$\mu$}{Conditional mean of $\fv_*$}
    \nomenclature{$\sigma^2$}{Conditional variance of $\fv_*$}
Here, $\Kxxs \in \mathbb{R}^{p \times n}$, so that $\mu(X_*)\in\mathbb{R}^{p\times 1}$ and $\sigma^2(X_*)\in\mathbb{R}^{p \times p}$. Equation \ref{eq:predict2} gives the expression of the conditional distribution of $\fv_*$ given test points $X_*$, and training data $X$ and $\fv$. The mean and variance of this predictive distribution are found via equations \ref{eq:mu} and \ref{eq:sigma}. In the context of Bayesian optimization, the surrogate model provides statistical information (i.e. mean and variance from equations \ref{eq:mu} and \ref{eq:sigma}) of how the underlying objective function $\fy$ behaves for all possible values ${\bf x}_*$ that have not yet been sampled. 

\subsubsection{The Acquisition Function: }\label{acq_fxn}
Generally, an acquisition function operates on the surrogate model and yields a function whose optimum reflects the most promising location for the next function evaluation to occur. More formally the acquisition function is defined as $a:(\bm x, \fs) \rightarrow \mathbb{R}$. In equation \ref{eq:acq_bo} the general surrogate model $\fs$ has been replaced with a Gaussian process to reflect the model used for \BO. To have more compact notation we hereafter refer to the acquisition function as $\fa$, which assumes the inclusion of the surrogate model as an argument. 
    \begin{align}
        \fa &\triangleq a(\bm x,\gp) \label{eq:acq_bo}
    \end{align}

\noindent Given an acquisition function, \BO{} selects $\hat{\bm x} = \argmax_{\mathcal{X}}\fa$ as the next location to be evaluated in the search process.

An important characteristic of any acquisition function is that it should not lead to greedy or myopic behavior. That is, $\fa$ should enable sufficient exploration and modeling of $\fy$ by sampling new locations, $\bm{x}$, that will improve $\fs$. At the same time, $\fa$ must efficiently exploit $\fs$ to reach the expected minimum of $\fy$ as quickly as possible. There are many ways to define $\fa$, although no single method is superior to all others in all applications (a manifestation of the `no free lunch' theorem)~\cite{Shahriari2015a,Brochu2010,Hoffman2011}. A few widely used methods are considered here: Expected Improvement (EI), Upper Confidence Bound (UCB), and Thompson Sampling (TS). 

    %\paragraph{Expected Improvement:} 
     The EI method seeks to select the next sample point to maximize the statistically expected improvement in the optimum when when the current best minimizer is $x^+$. 
     The EI function (see~\cite{Jones1998}) is defined by %%equations \ref{eq:EI1} and \ref{eq:EI2}.
    \begin{equation}
        a_{\text{EI}}(\bm x) = \begin{cases}
                            (\mu(\bm{x}) - \fv(\bm{x}^{+}))\Phi(Z) + \sigma(\bm{x})\phi(Z) &,  \quad \sigma(\bm{x})>0 \\
                            0 &, \quad \sigma(\bm{x})=0
                        \end{cases} \label{eq:EI1}        
    \end{equation}
    \begin{equation}
        Z = \frac{\mu\left( \bm{x} \right) - \fv(\bm{x}^{+})}{\sigma(\bm{x})}, \label{eq:EI2}
    \end{equation}
where $\mu(\bm{x})$ is the mean predicted value of the GP at $\bm{x}$ and $\sigma(\bm{x})$ is the predicted standard deviation at $\bm{x}$, $\fv(\bm{x}^{+})$ is the best observed value of the objective function\footnote{This should be substituted by the best mean prediction if the objective function is stochastic}, and $\Phi(Z)$ and $\phi(Z)$ are the PDF and CDF of the standard normal distribution $Z$, respectively. 

    %\paragraph{Upper Confidence Bound:} 
%%\nisarcomm{UCB should be LCB, since we are defining optimization to be minimization problem??}		
The UCB acquisition function was originally inspired by the UCB regret in multi-armed bandit problems~\cite{Srinivas2010}. The key idea is to bound regret in a sequential optimization process, where regret is defined as the difference between the actual strategy (based on decisions made with imperfect information) and the ideal strategy (decisions made with perfect information). 
UCB is defined as 
    \begin{align}
        a_{\text{UCB}}(\bm{x}) &= \mu(\bm{x})+\beta \sigma(\bm{x}) \label{eq:UCB}
    \end{align}
where $\mu(\bm{x})$ and $\sigma(\bm{x})$ again come from the GP, and $\beta>0$ is a hyperparameter. UCB thus simply takes the mean GP prediction and adds some multiple $\beta$ of the standard deviation to the mean GP prediction for every $\bm{x}$ (note: in minimization problems, the lower confidence bound (LCB) obtained by subtracting $\beta \sigma(\bm{x})$ is used instead -- to remain consistent with the predominant terminology in the \BO{} literature, however, this acquisition function will still be referred to as the UCB for the remainder of this paper). 
One drawback for this function is that $\beta$ must be tuned; while principled ways exist to do this (e.g.~\cite{Shahriari2015a}), alternatives are often sought to avoid introducing this extra dependence. %%(as well as the fact that UCB has not proven to be consistently better than other methods such as EI, which require no hyperparameters \cite{Snoek2012}).

    %\paragraph{Thompson Sampling:} 
Thompson sampling (TS) involves drawing functions directly from the GP surrogate. In practice, sampling a function from eq. \ref{eq:predict2} means calculating the predicted joint GP mean and covariance at a finite set of points in $\bm{x}_*$, and then sequentially drawing $\fv_*$ values at these points based on the resulting conditional GP means and variances. Once a function sample is obtained, its optimum is easily found and the objective function is evaluated at that point. Due to its stochastic nature, TS naturally incorporates explore-exploit behavior. TS also naturally lends itself to parallelized searches and (from the law of large numbers) will tend to identify the most likely locations for $\bm{x}^*$ for sufficiently large samples sizes. However, since the complexity of GP inference is cubic in $n$, sampling GPs with sufficient resolution for $d\geq 3$ is usually quite expensive ~\cite{Shahriari2015a}.

\subsubsection{Bayesian Optimization Procedure and Example} 
    Algorithm \ref{alg:BayesOpt}, gives the \BO{} procedure. The termination criteria could be based on iteration thresholds, tolerances on changes to the optimum $\bm{x}$ and/or $y$ between iterations, or other methods ~\cite{Johnson}. In practice, the performance of Bayesian optimization depends greatly on the selection and parameterization of the surrogate model, along with the number and placement of the initial training observations (i.e. seed points). It is also important to consider how samples are drawn with respect to the acquisition function. These issues are addressed in more detail for the aerial combat simulation application in Section \ref{sec:methodology}.

        \begin{algorithm}[!htbp]
            % Bayesian optimization procedure
            \caption{\BO{}}
            \label{alg:BayesOpt}
            \begin{algorithmic}[1] % The number tells where the line numbering should start
                    \State Initialize surrogate model with seed data $\fv$ and hyperparameters
                    \While {termination criteria not met}
                        \State ${\bm x}_j = \argmax_{\mathcal{X}} \fa$ %\Comment equation \ref{eq:EI1} when applying the EI criterion
                        \State Evaluate $y(\bm{x}_j)$
                        \State Add $y(\bm{x}_j)$ to $\fv$, $\bm{x}_j$ to $X$, and update hyperparameters \label{ln:testing}%\Comment see Sec. \ref{sec:MLE_MAP_CV} 
                    \EndWhile
            \end{algorithmic}
        \end{algorithm}

    The statement on line \ref{ln:testing} merits extra attention. Since \BO{} is an iterative process, newly sampled data $\bm{x}_i$ and $y(\bm{x}_i)$ are added to the GP surrogate model each iteration. The parameters of the covariance kernel are recalculated by methods discussed in section \ref{sec:MLE_MAP_CV}. As a result, the acquisition function changes from one iteration to the next (sometimes drastically); this behavior will be highlighted in the following example.

%\subsubsection{Bayesian Optimization Example} \label{sec:BayesOpt_example}
\begin{figure}[!htbp]
      \centering
      \includegraphics[width=1.0\textwidth]{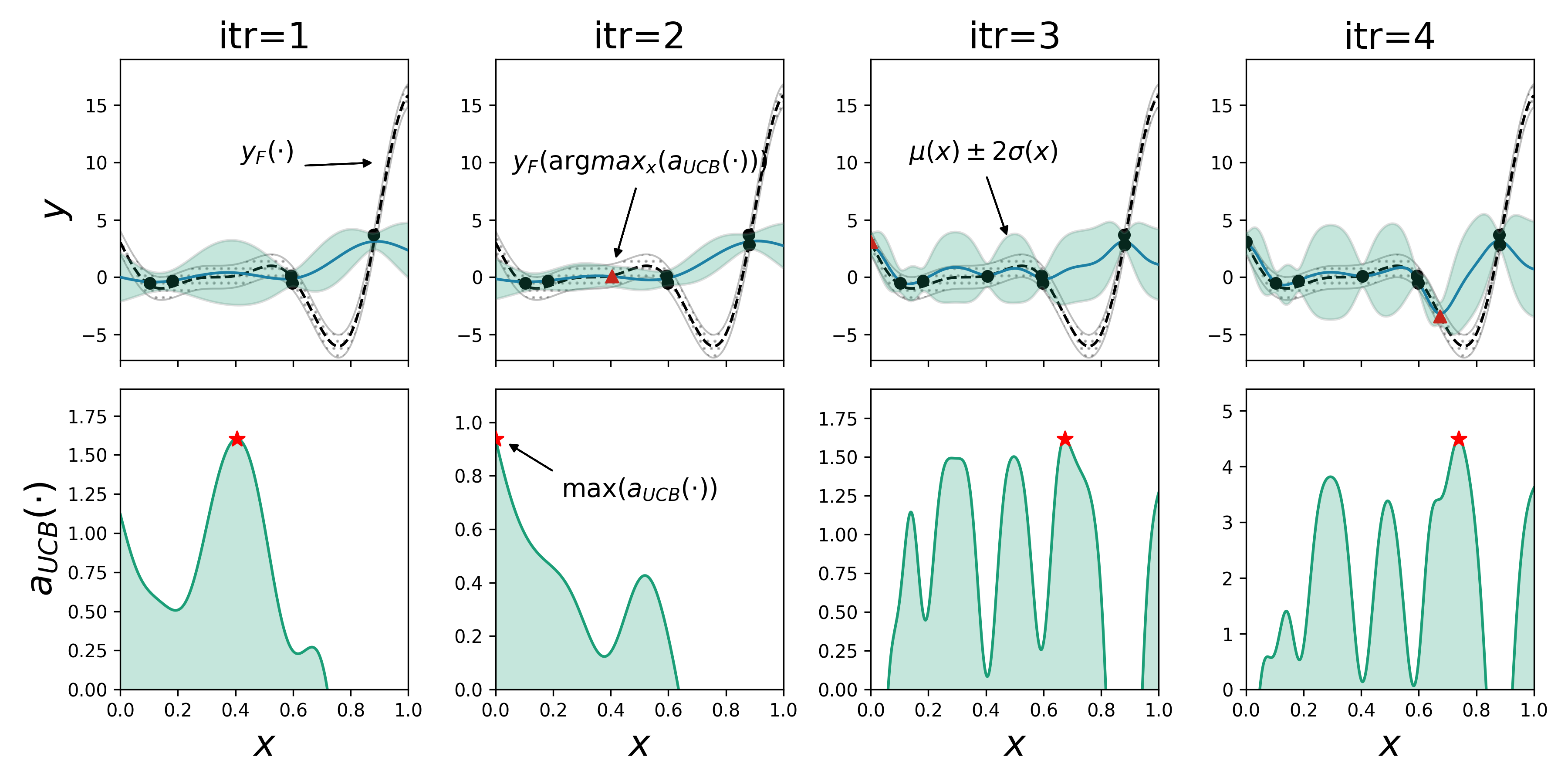}
      \caption{\BO{} example on the Forrester function using the UCB acquisition function. Red triangles are new data added to the GP, red stars are the maximum of the acquisition function where the next evaluation of the objective function will occur. }
      \label{fig:UCB_example}
\end{figure}

Figure \ref{fig:UCB_example} shows a few iterations of the \BO{} process for an illustrative toy example based on the 1D Forrester function, 
\begin{align}
    y_{F}&= (6x-2)^2\sin(12x-4) \quad \text{for } x=[0,1] \label{eq:forrester}
\end{align}
which is a canonical optimization objective function~\cite{forrester2008engineering} with multiple local maxima, multiple local minima, and a saddle point. In this example, Gaussian noise with zero mean and standard deviation equal to 0.5 has been added to $y_F$ (indicated by the lightly shaded gray region). 

The columns in Fig. \ref{fig:UCB_example} show successive iterations of the \BO{} algorithm; the top and bottom rows show $\fs$ and $\fa$, respectively. In this case $n=6$ seed points have been randomly selected for the GP surrogate $\fs$. These points are shown in the top left plot, along with the associated GP predictions. The values from the GP are obtained using eq. \ref{eq:predict2} to evaluate a fixed mesh of points over the range of $\bm{x}$. Note that with only six seed data points, the GP does not represent $y_{F}$ very well. 

The second row of Fig. \ref{fig:UCB_example} shows the acquisition function $\fa$, which in this example is given by the UCB. On the first iteration, $\fa$ is maximized at approximately $\bm{x}=0.4$ (red star); $y_{F}$ is then evaluated at this location on the second iteration of \BO{}, and a new data point is added to the GP (red triangle), so that the sampling-update cycle repeats. Also notice that the GP uncertainty increases and decreases in certain areas from iteration to iteration; this is due to the fact that the hyperparameters governing the covariance function (equation \ref{eq:gp_cov}) are updated when more data is acquired (in particular, the GP tends to become less uncertain in areas it has sampled already). 
In the four iterations shown here, although the GP does not represent $y_{F}$ very accurately, it does manage to approximate $y_F$ accurately in regions of interest, i.e. where the minimum might actually exist. On the fourth iteration, $y_F$ is evaluated near the true minimizer and another point nearby this minimum will be selected via $\fa$ for evaluation on the fifth iteration.

This example highlights several key takeaways. Firstly, the number and location of seed points affects the initial representation of the surrogate function and acquisition function. Secondly, the GP attempts to estimate the true objective function, but only has information given by previous experiments. Thirdly, the GP changes during optimization when trying to re-learn the kernel hyperparameters with the newly acquired sample data. Finally, $\fa$ will change when the GP surrogate $\fs$ changes. This means that if the GP is a poor model of $\fy$ and does not accurately represent uncertainty in unsampled parts of $\mathcal{X}$, then the \BO{} process will tend to perform poorly. An ideal sampling strategy would yield a GP that provides a good global model of $\fy$ with as few evaluation points as possible. 

% \nomenclature{HRMS}{Hybrid Repeat/Multi-point Sampling}
% \nomenclature{$F_{f(\cdot)}$}{Forrester objective function}

    \section{Bayesian Optimization for Decision-Making AI Adaptation}\label{sec:methodology}
Bayesian optimization has emerged as a critical tool for `auto-tuning' various machine learning and AI algorithms~\cite{Bergstra2011,Snoek2012,Mahendran2012}, and has been applied to modeling of user preferences and reinforcement learning~\cite{Brochu2010}. It is well suited for optimizing bounded objective functions that are unknown and expensive to evaluate, and is often able to do this with the fewest function evaluations as compared to other competing methods~\cite{Jones1998}. These properties make it appealing for optimally adapting the behavioral parameters of an AI decision maker. In this section, the overall \BO{} learning process for the aerial combat simulation application is described first. Then, specific strategies for coping with the major practical implementation issues for \BO{} are described.  

%\nisarcomm{EDIT THIS TO REFLECT NEW FACTORIZATION: } 
%However, as mentioned earlier, standard \BO{} techniques do not perform well on objective functions that are highly volatile and when accurate models of the true objective function are desired -- as in the case of AI decision-maker tuning for the aerial combat simulation application considered here. To this end, we propose and demonstrate a novel sampling approach called Hybrid Repeat/Multi-point Sampling (HRMS). HRMS not only yields more repeatable optimization results, but also provides surrogate functions that more accurately represent the true objective surface when the optimization is completed. 
\subsection{Learning Engine} \label{sec:LearningEngine}
\begin{figure}[!htbp]
          %Problem 3 Figure
          \centering
          \includegraphics[width=0.5\textwidth]{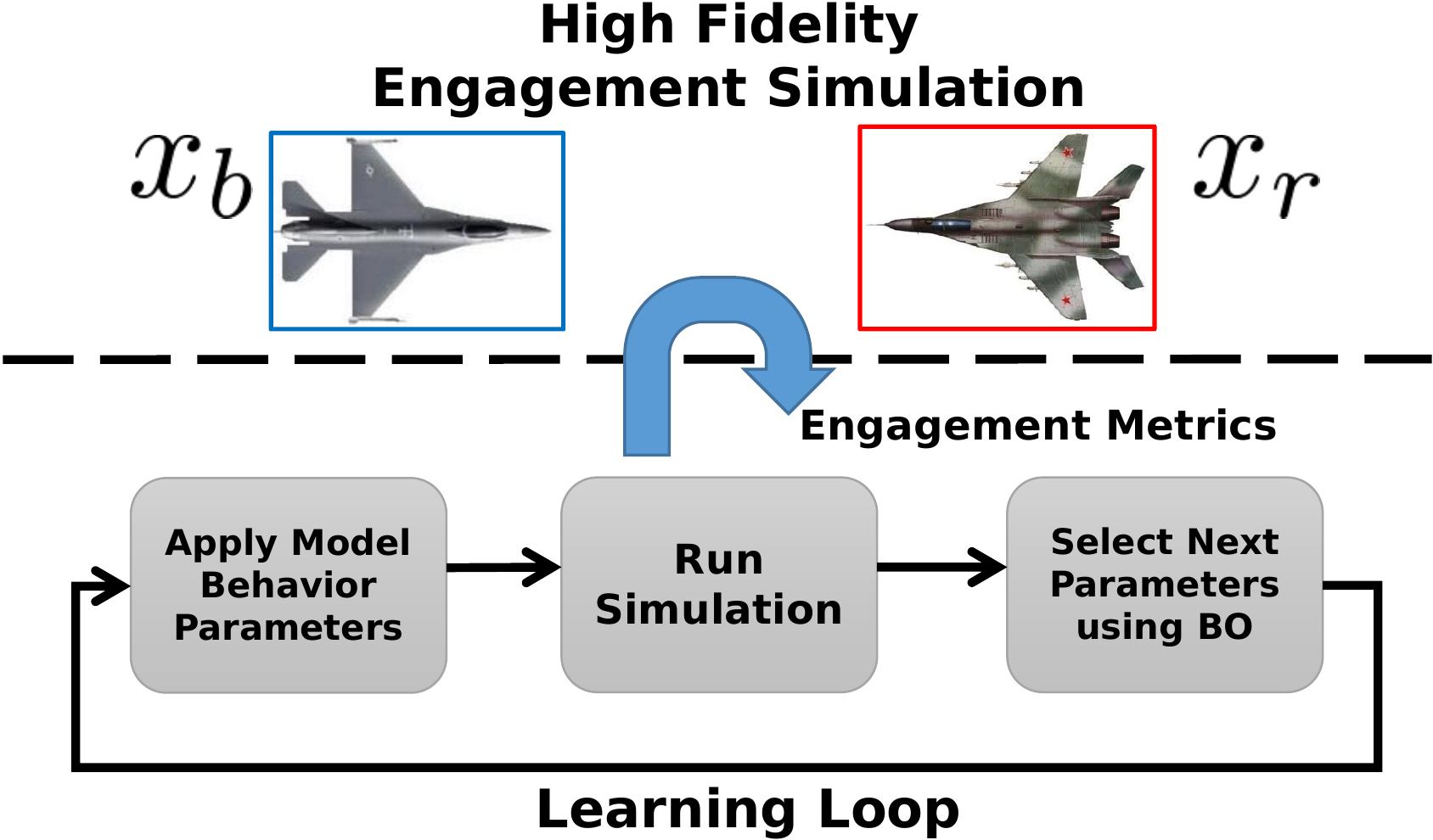}
          \caption{Learning Engine Diagram. Representation of the engagement simulation environment (top) and the high-level learning loop (bottom)}
          \label{fig:LearningEngine}
    \end{figure}

For simplicity, the focus hereafter will be on adapting blue's behavioral parameters only for simulated AI vs. AI combat. 
As such, the red agent is treated as an AI agent with constant default parameter values $\bm{x}_r$ (although the learning procedure does not require this assumption). 
Figure \ref{fig:LearningEngine} depicts the process by which the optimum blue parameters $\bm{x}_b$ are learned. 
The bottom portion of the figure shows the high-level learning loop, while the top of the figure represents the simulation environment. An initial set of agent parameters is chosen first, and then a simulation is run using the parameters for each agent using the Mission Simulation System (MSS) simulation environment developed by Orbit Logic, Inc. for Live-Virtual-Constructive (LVC) applications. MSS employs very high fidelity models of aircraft, their avionics system, installed sensors, weapons \& countermeasures, and external situation sensors. Additionally, AI agent models were developed to support tactics development, war-gaming and pilot training. The AI agents allow complex blue/red engagements to be played out without any humans in the loop, or with a mix of agents and humans. 
Table \ref{tab:agent_params} in Sec. \ref{sec:problem_definition} shows the behavioral parameters for the proprietary pilot agent AI developed by Orbit Logic. 
The overall learning approach is agnostic to the internal details of the agent's AI, and thus any other AI with tunable behavior parameters could be used instead. 

At the end of a simulated engagement, one of the summary engagement metrics $y_m(\bm{x}_r,\bm{x}_b)$ shown in Table \ref{tab:objective_metrics}, such as TTK, energy management, number of objectives completed, etcetera is calculated based on recorded data for the red and blue aircraft, and then delivered to the \BO{} component. The \BO{} algorithm selects one or more $\bm{x}_b$ values to implement for the next round of engagement simulations that will potentially optimize the desired $y_m(\bm{x}_r,\bm{x}_b)$. This initiates the simulation-learning-parameter update cycle again; if more than one $\bm{x}_b$ value is selected by \BO{}, then each value results in a separate parallel instance of a follow-up simulated engagement. The cycle repeats until the desired termination condition is met. Possible termination criteria include: (i) no significant changes in either $\bm{x}_b$ or $y_m(\bm{x}_r,\bm{x}_b)$ are found; (ii) the maximum number of optimization iterations has been reached; (iii) the maximum number of simulated engagements has been reached. Note that (ii) and (iii) are identical if only one $\bm{x}_b$ value is selected by \BO{} following a single engagement simulation instance. 

\subsection{Engagement Simulation Setup} \label{sec:SimulationSetup}
    A one-on-one engagement scenario was designed to collect relevant metrics for training and validating the \BO{} tuning process. 
    For ease of explanation and analysis, all results and discussion for \BO{} in the air combat simulation application will be based on this single scenario. 
    The scenario includes a single blue force fighter jet penetrating an adversary's defended engagement zone. The simulation is run for $T_{max}=300$ seconds, or until either the blue or the red agent is eliminated. The blue agent's primary objective is to engage in air-to-air combat against a red defending fighter jet to achieve theater control. The blue agent's `goal' is to minimize the TTK, which is defined in equation \ref{eq:TTK}. Thus, $TTK>300$ represents blue being eliminated, $TTK<300$ represents blue victory, and $TTK=300$ denotes that both fighters survived.

	\begin{align}
		TTK = \begin{cases}
		 		SimTime &\quad\text{, if Blue survived}\\
		 		2*T_{max} - SimTime &\quad\text{, otherwise}
			  \end{cases}
		\label{eq:TTK}
	\end{align}

\begin{figure}[!htbp]
          %Problem 3 Figure
          \centering
          \includegraphics[width=0.75\textwidth]{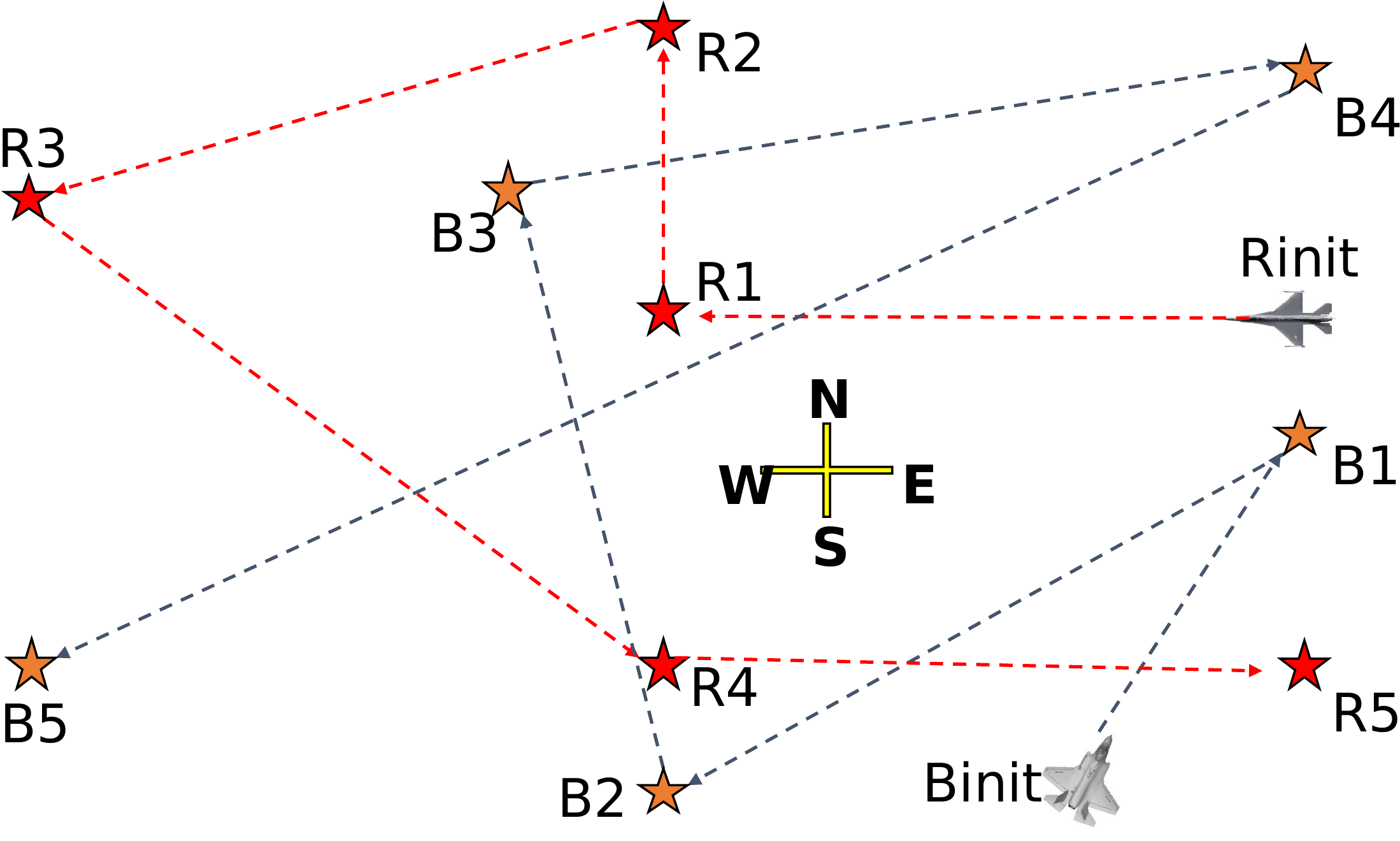}
          \caption{Diagram depicting the engagement space. Red and blue players begin at $R_{init}$ and $B_{init}$ respectively. Their flight path carries them over the way-points indicated by the dashed lines. During the flight they will engage if criteria in their decision logic are met.}
          \label{fig:s1}
    \end{figure}

    Each agent is assigned a nominal flight plan that includes multiple way-points that, when flown over, provide mission `scoring points'. The way-points are arranged in such a way as to ensure that the fighters periodically encounter each other as illustrated in Figure \ref{fig:s1}, instigating the employment of air-to-air combat logic. Engagement metrics are calculated at the conclusion of the simulation and then delivered to the Learning Engine.

The remaining subsections describe specific practical issues that arise for running the \BO{} learning process, and the solutions developed to address these. 

\subsection{Initial Data Seeding}\label{sec:seeding}
%A subtle but critical point is that, for \BO{} to work, the surrogate function should accurately approximate the true objective function. 
%\nisarcomm{Don't really need to mention this all to motivate why seed samplng is necessary: we simply have to start the BO process somewhere since a GP is only properly instantiated as a stochastic process realization if we observe it at certain points. This remaining part of this paragraph doesn't necessarily follow from this first sentence -- it is also a bit of a contentious point: a perfect surrogate model leads to good optimization, but good optimization does not require a perfect model if it captures the relevant features...this should be expounded upon a bit more carefully elsewhere.}   
For initialization, it is necessary to to bootstrap the GP surrogate model with an adequate `seed' data sample. To get a good seed sample of the $\bm{x}_b$ variable space , random sampling is often used~\cite{Bergstra2012}. However, to ensure sufficient coverage of the $\bm{x}_b$ space, na\"{i}ve uniform random sampling is inefficient, as the samples are not correlated to one another and thus can be undesirably drawn in proximity to each other. 
Instead, a type of stratified quasi-random sampling called Latin hypercube (LH) sampling is used~\cite{Macdonald2009,Schonlau1997} to ensure a more uniform initial sampling of $\bm{x}_b$.  %%Unfortunately, the fact that the objective function is unknown limits the ability to optimally select the number of seed points.

The success of \BO{} rests heavily on the GP's ability to globally model the true objective function. 
Without proper initial sample seeding, the \BO{} algorithm can easily get trapped by local minima of the objective function and not escape. 
The number of random initial seed points to use mainly depends on the homogeneity of the objective function (i.e. whether it behaves similarly for different parts of the $\bm{x}_b$ space) and the smoothness of the function. 
To the best of our knowledge, quasi-random sampling with a heuristically selected amount of seed data is currently state-of-the-art for seeding ~\cite{Hoffman2014,Chevalier2013}. It is common to use $n=10d$ seed points, where $d$ is the dimension of $\mathcal{X}$~\cite{Jones1998,Huang2005}. %In section \ref{sec:parallel_search}, we suggest that, depending on the properties of the objective function, it is necessary to use a different method other than a simple multiple of the dimensionality. \nisarcomm{This isn't really brought out in that section, is it?}

\subsection{Selecting sample points from the acquisition function}\label{sec:candidate_selection}
The first step  of the \BO{} loop in Algorithm \ref{alg:BayesOpt} is to find the $\argmax$ of the acquisition function $\fa$. However, this is not a trivial matter. As shown in Fig. \ref{fig:UCB_example}, %and \ref{fig:1dBO_example}). 
it is very common for $\fa$ to have many local maxima, regardless of which flavor of acquisition function is used. 
It is thus necessary to apply a global non-linear/non-convex optimization algorithm for this step. 
Although this means that another optimization algorithm must be run within \BO{},  
it is important to recall that this is in fact perfectly acceptable, since the true objective function $\fym$ is highly volatile and expensive to optimize using a global non-linear/non-convex optimization algorithm. 
As such, it is computationally far more efficient and operationally more acceptable/desirable to use such an optimization algorithm on $\fa$. 

This work makes use of the DIRECT (Dividing rectangles) optimization algorithm to find the best local maximum of $\fa$. DIRECT is also widely used elsewhere in the \BO{} literature \cite{Ponweiser2008,Hoffman2011,Mahendran2012,Wang2013}. This method uses the Lipschitz continuity properties of $\fs$ to bound function values in local rectangles and search accordingly~\cite{Jones1993}. Further details on DIRECT and a discussion of other possible global optimization algorithms for \BO{} can be found in \cite{Shahriari2015a}.

In standard \BO, only a single new $\bm{x}_b$ point is sampled according to $\fa$ on each optimization iteration. However, as discussed in Sec. \ref{sec:parallel_search}, it is also possible (and sometimes much more beneficial) to redefine $\fa$ so that multiple $\bm{x}_b$ points are sampled simultaneously on each iteration. 

%%%%  			
\subsection{Updating the kernel hyperparameters}\label{sec:MLE_MAP_CV}
The kernel used to define $k$ in this research is the Mat\'{e}rn kernel,
\begin{align}
k_{\nu = 3/2}\left( \bm{x}_{b,i}, \bm{x}_{b,j} \right) &= \sigma_0\left( 1 + \frac{\sqrt{3} r_{ij}}{\ell} \right)\exp\left( - \frac{\sqrt{3}r_{ij}}{\ell} \right), \\
\label{eq:Mat32}
r_{ij} &= \sqrt{(\bm{x}_{b,i} - \bm{x}_{b,j})^T(\bm{x}_{b,i} - \bm{x}_{b,j})},
\end{align}
with hyperparameters $\sigma_0$ and $\ell $, which are the kernel amplitude and length-scale, respectively. 
The Mat\'{e}rn kernel is widely used in \BO{}, since it has the useful property that different values of $\nu$ yield a kernel with different degrees of differentiability (where $\nu$ is nearly always taken to be half integer to simplify the kernel expression).    
The Mat\'{e}rn kernel is guaranteed to be $k$ times differentiable when $k\leq \nu$. Therefore the kernel with $\nu=3/2$ is $k=1$ times differentiable and reflects the prior belief that the outcome engagement metrics $y_i$ for the aerial combat simulations do not have higher order derivatives. 
Note that eq. (\ref{eq:Mat32}) is the `isotropic' version of the Mat\'{e}rn kernel, where $\ell$ is scalar and each dimension of $r_{ij}$ has equal influence. 
Incorporating separate $\ell$ terms for each element of $r_{ij}$ leads to the non-isotropic Automatic Relevance Determination (ARD) version of the kernel.  
An ARD kernel can automatically learn to ignore parts of $\bm{x}_b$ that are irrelevant to predicting the output~\cite{Rasmussen2006}, by scaling variables based on their contribution to the covariance function eq. (\ref{eq:gp_cov}). 
However, ARD kernels lead to more hyperparameters than non-ARD kernels and thus can overfit data more easily. 
As is standard in GP regression, an additive observation noise variance $\sigma_n^2$ is also assumed for each training datum $f(\bm{x}_{b,i})$, where
\begin{align}
f(\bm{x}_{b,i}) &= y(\bm{x}_{r},\bm{x}_{b,i}) + \epsilon_i, \label{eq:noisy_var}\\ 
\epsilon_i &\sim {\cal N}(0, \sigma^2_n), \label{eq:noise}
\end{align}
Hence, the full set of hyperparameters $\Theta = \left\{\sigma_n^2, \sigma_0, \ell \right\}$ governs the behavior of the covariance function of the GP in eq. (\ref{eq:gp_cov}). 

Since the best $\Theta$ values are unknown a priori, they must be learned and updated during \BO{}. Point estimation strategies based on maximum likelihood estimation (MLE) are the most widely used in the \BO{} literature for supervised learning of $\Theta$. 
%    \paragraph{Maximum likelihood estimate (MLE): }\label{MLE}
MLE maximizes the GP likelihood function $p(\fv|X,\Theta,k)$ with respect to $\Theta$, where
\begin{align}
			p(\fv|X,\Theta,k)   &= {\cal N}(\mu(X),K_{\fv}(\Theta,k)) \nonumber \\
							&= (2\pi)^{-D/2}|K_{\fv}(\Theta,k)|^{-1/2}\exp \left(-\frac{1}{2}(\fv-\mu(X))^T K_{\fv}(\Theta,k)^{-1}(\fv-\mu(X)) \right), 
							\label{eq:gp_datalik} \\
			K_{\fv}(\Theta,k) &= K(X) + \sigma_n^2 I.  
		\end{align}
Fast gradient-based convex optimization techniques are most commonly used to minimize the negative logarithm of eq. (\ref{eq:gp_datalik}) (i.e. the negative log likelihood), since the required derivatives can be obtained analytically. 
However, since the GP likelihood is generally non-convex, numerical optimization can converge to many different local optima for $\Theta$. 
To further complicate matters, the best local optimum may be undesirable for learning with sparse data early on in the \BO{} process, since the associated $\Theta$ values typically lead to overfitting of the training data~\cite{Cawley2007, Rasmussen2006}. 
This behavior is especially important to consider when trying to minimize the number of simulations for \BO{}. %and when the GP surrogate model has many hyperparameters. 
%    \paragraph{Cross-Validation: }\label{CV}
Cross-validation (CV) methods are commonly used with MLE to avoid overfitting. 
In CV, the GP is trained on a designated training data set using MLE, and then validated by assessing the predictive error on a completely independent test data set, to ensure that the learned model is not overfit. 
Although unbiased, CV tends to produce undesirably high error variances for small data sets~\cite{Cawley2010}, making it difficult to reliably learn $\Theta$. %Hence, it is difficult to use CV to reliably identify $\Theta$  with small initial data sets in \BO{}. 

%    \paragraph{Maximum a posteriori (MAP): }\label{MAP}
Alternatively, maximum a posteriori (MAP) estimation can be used to avoid overfitting by adding hyperpriors $p(\Theta| k)$ on $\Theta$. In this case, the posterior distribution for $\Theta$ from Bayes' rule
    \begin{align}
        p(\Theta| \fv,X,k) &\propto p(\fv| X,\Theta,k) p(\Theta| k) \label{eq:MAP}
    \end{align}
is maximized via gradient ascent. 
Considering the logarithm of eq. (\ref{eq:MAP}), the $p(\Theta| k)$ hyperpriors act as additional `regularization' variables for the GP data log-likelihood function in eq. (\ref{eq:gp_datalik}). This can help smooth away poor local optima, and prevent point estimates of $\Theta$ from overfitting or changing too drastically given small amounts of data. 
As the number of sample points added to the GP increases, the contribution of $p(\Theta| k)$ to the point estimate of $\Theta$ diminishes. 
In this work, $p(\Theta| k)$ hyperpriors are simply defined as uniform distributions for a broad range of $\Theta$ values. 
Following MAP updates to the hyperparameters, any predictions made through the surrogate function can be modified to account for hyperparameter uncertainty by integrating with respect to $\Theta$, i.e. by marginalizing out the hyperparameters,
    \begin{align}
        p(\fv_*| X_*, \fv, X, k) &\propto \int_{-\infty}^{\infty}p(\fv_*|X_*, \fv, X, \Theta,k) p(\Theta|\fv, X, k) d\Theta \label{eq:MAPpredict}.
    \end{align}
Since both $p(\Theta|\fv,X,k)$ and this integral are in general analytically intractable, approximation schemes like the Laplace method can be used. 
In the Laplace method, $p(\Theta|\fv,X,k)$ is approximated as a multivariate normal ${\cal N}(\hat{\Theta}_{MAP},\hat{\Sigma})$, where $\hat{\Theta}_{MAP}$ is the MAP estimate and $\hat{\Sigma}$ is the negative inverse Hessian of the RHS of eq. (\ref{eq:MAP}) evaluated at $\hat{\Theta}_{MAP}$. Eq. (\ref{eq:MAPpredict}) then becomes analytically tractable, so that it can be used in the acquisition function. This approach effectively inflates the prediction uncertainty of the GP by the uncertainty in $\Theta$, which is especially useful for the initial phases of \BO{} when sparse data is available.

The specification of hyperpriors is an excellent place to build in expert knowledge of the system if it is available. For example one might tightly constrain the hyperparameter governing the length scale, if there were some reason to believe that the length scale should be near a certain value. In the air combat scenario, where we did not have much a priori knowledge, we used uniform, weak, prior distributions to allow the observations to have more effect on the learning process.

\section{Serial, Parallel and Hybrid Sampling Strategies}\label{sec:parallel_search}
One of the strengths of \BO{} is that it can greatly reduce the number of required evaluations of an expensive objective function, which in this case requires a full simulation of an aerial engagement. In fact, \BO{} is often able to find the optimum in fewer iterations than any other known optimization method~\cite{Jones1998}. However, if multiple new simulations can be specified and possibly run in parallel, the optimization process can be greatly accelerated and improved in terms of reliability and repeatability (a very desirable characteristic for black-box optimization with noisy objective functions). 
In the context of air combat simulation training, such parallelization is especially attractive for adapting AI vs. AI engagements, as well as developing aggregated pilot performance evaluations (i.e. where instances of the same AI are simultaneously pitted one-on-one against different human-controlled red agents in different simulation instances). 

Three approaches can be considered for parallel sampling-based search. 
The first approach is to use $\fa$ to identify several different $\bm{x}_b$ locations simultaneously, e.g. the best local optimum and other `nearly optimal' locations simultaneously, or a sequence of hypothetically optimal locations. This will be referred to as \textit{multi-point sampling} (MS)\footnote{also known as batch sampling}. 
The second is to sample the same maximum of $\fa$ multiple times. This method will be referred to as \textit{repeat sampling }(RS).  
The traditional, non-parallel approach to sampling new $\bm{x}_b$ via maximization of $\fa$ is denoted \textit{single sampling} (SS), which is a special case where RS=MS=1.  

\begin{figure}[t]
            % \centering
            \captionsetup[subfigure]{farskip=-0.15cm,captionskip=-0.5cm,labelfont={large,bf}}
            \subfloat[]{%
                \includegraphics[width=1\linewidth]{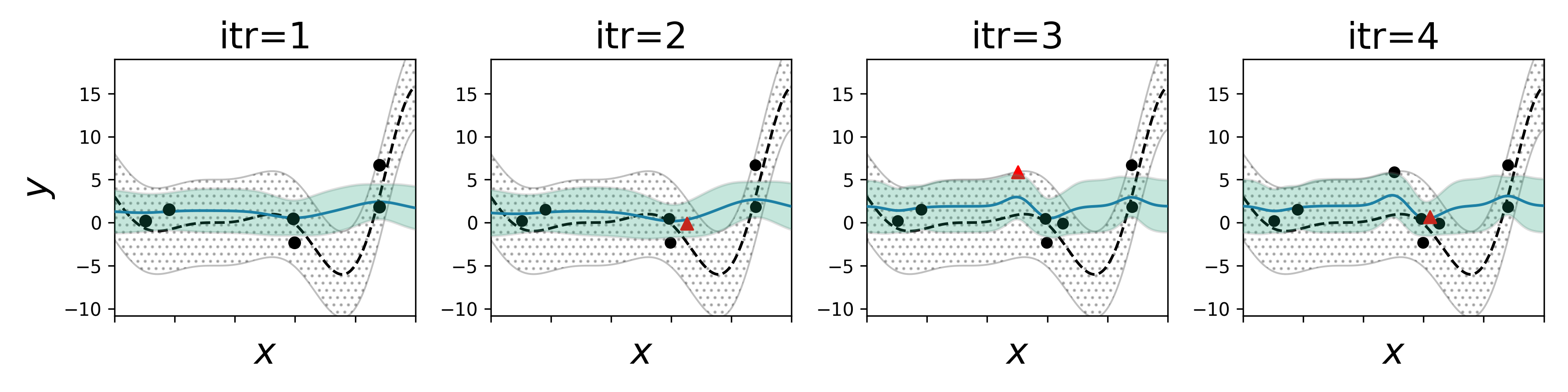}%
                \label{fig:SSForrester}%
            }\\
            \subfloat[]{%
                \includegraphics[width=1\linewidth]{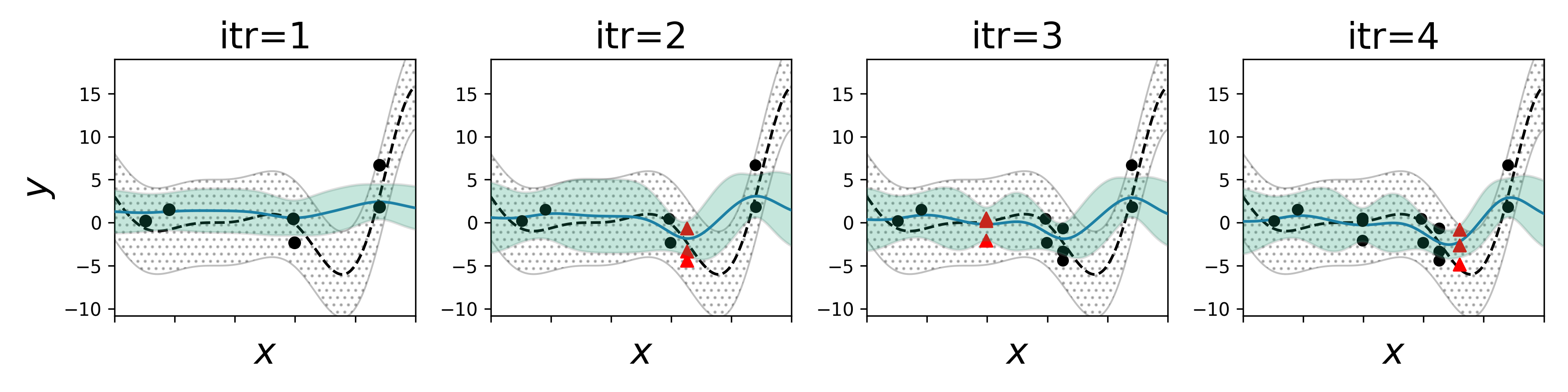}%
                \label{fig:RS3Forrester}%
            }\\
            \subfloat[]{%
                \includegraphics[width=1\linewidth]{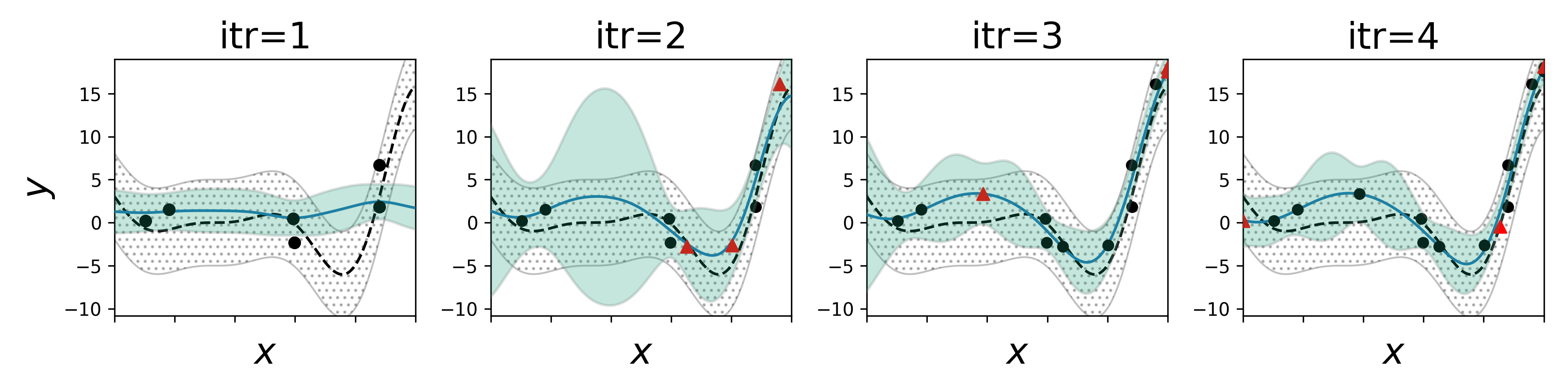}%
                \label{fig:MS3Forrester}%
            }%
            \caption{Examples of SS, RS and MS sampling on 1D Forrester function (red triangles indicate new samples): 
						(a) SS with UCB acquisition function, 
						(b) RS=3 and MS=1 using UCB, (c) RS=1 and MS=3 using GP-UCB-PE. }
            \label{fig:RSvsMS}
        \end{figure}

Figure \ref{fig:RSvsMS} illustrates the SS, RS and MS sampling methods on a much noisier version of the 1D Forrester function shown earlier, using the parallel sampling acquisition functions described below for each approach. 
From this simple example, it is easy to see that (in just 4 optimization iterations) the resulting GP surrogate models for both the RS and MS methods do a much better job at estimating the statistics of the underlying objective function than the traditional SS approach. 
Clearly, this improvement is due in large part to the fact that both MS and RS benefit from additional function evaluations and thus can use more data in their GPs to make better predictions.  
However, note that the MS approach at iteration 2 is already significantly better than SS at iteration 4, i.e. even after the same number of new data samples have been added to both GPs. 
Hence, strategic placement of additional samples earlier on in the \BO{} process can yield significant benefits. 
As shown later in Sec. \ref{sec:results}, the improved performance for RS and MS over SS (given the same total number of function evaluations) also generally appears to hold in more complex problems.  
With this in mind, different techniques for implementing MS and RS methods are considered next. 

\nomenclature{RS}{Repeat Sampling}
\nomenclature{MS}{Multi-point (or Batch) Sampling}

\subsection{Multi-point sampling}
       Multi-point sampling (MS) uses any one of several strategies to forecast which points might be of the most interest in future evaluations ~\cite{Snoek2012,Chevalier2013,Contal2013}. Three different methods are considered here. 
        
        The first $q$-EI method, introduced by \cite{Schonlau1997,Ginsbourger2010}, is similar to EI for traditional SS -- except that, instead of optimizing over $\bm{x_b}$, the EI function is optimized over multiple ($q>1$) sample locations $\bm{x}_b$  at once. This method is notoriously expensive to evaluate, since the dimensionality of $\fa$ increases (e.g. with $q=5$ and $d=2$ leads to a 10 dimensional acquisition function), thus causing both calculation and optimization to become much more expensive.  
        %It is so expensive that brute force Monte Carlo simulations are known to be faster in some cases. %--be more precise: faster in what sense?
As an alternative, the approximation method introduced by \cite{Chevalier2013} is often used for efficient computation of the $q$ points. Their result bypasses the need for Monte Carlo simulations by applying the so-called ``Tallis Formula'', which uses a tractable method for calculating the conditional expectations of the multinormal distribution.

%\nisarcomm{Again: Is this LCB or UCB since we do minimization?}
The second GP-UCB-PE method ~\cite{Contal2013} uses the UCB function and a pure exploration (PE) technique to perform parallel search for the optimum. This approach is based on a simple premise: greedily select $q$ points in sequence using the UCB acquisition function. This is done by first selecting the maximum of the UCB function as $\bm{x}_1$, exactly as in a conventional SS approach. Next, taking advantage of the fact that the surrogate model conditional covariance predictions in eq. (\ref{eq:gp_cov}) are independent of the exact value of $\fv$, the UCB can be recalculated assuming the proposed $\bm{x}_1$ is given as part of the covariance function, which allows for selection of $\bm{x}_2$ at the next UCB maximum. This process is repeated until $q$ points have been selected. 

Finally, a simple extension of basic Thompson sampling can also be used for MS. Instead of drawing a single function from the GP surrogate model and selecting its optimum, $q$ functions are drawn together at once. Then the optimum of each sample function is found, and experiments are run at each of the corresponding locations of an optimum. As in the case of Thompson sampling for classical SS, this approach is asymptotically efficient but still suffers from the curse of dimensionality. %: optimization of random functions in high-dimensional spaces requires that $\bm{x}_b$ be randomly sampled with a sufficiently fine resolution, which is computationally demanding. 

\subsection{Repeat Sampling} 
Repeat Sampling (RS) has not been previously studied in the \BO{} literature, and is thus a novel contribution of this work. It is argued here that RS can be especially useful in problems with very noisy and volatile objective functions, i.e. like those found in simulated aerial combat. The rationale behind using RS is intuitively related to the concept of `repeated measures' in experimental design~\cite{pyzdek2003quality}. Loosely speaking, if the variance for the outcome of any given individual experiment (simulation) is large, then repeating the experiment at the same setting for the independent variables ($\bm{x}_b$, in this case) allows statistical inference to be made with fewer experiments by controlling the variance and reducing sensitivity to outliers.

The connection to the mechanics of \BO{} stems from the kernel function hyperparameter updates at the end of each optimization iteration. As discussed previously, the hyperparameter update requires solving a non-convex optimization problem involving the GP data log-likelihood. If only a single noisy sample is evaluated in an iteration, the kernel parameter update could respond poorly to noise or outliers, and thus converge to a poor local optimum in $\Theta$ space and/or produce an incorrect surrogate GP that overfits data. The resulting GP model $\fs$ will then not accurately reflect the true statistics of $\fym$, which in turn will corrupt $\fa$ and can subsequently lead to suboptimal selections of $\hat{\bm{x}}_b$ for all remaining \BO{} iterations. This issue is especially critical for the first few iterations of \BO{}, since the GP surrogate will be based on relatively small data sets. However, if a larger `statstically representative' sample of $\fym$ is obtained for any given $\bm{x}_b$ using RS in such cases, it becomes possible to reduce the sensitivity of the hyperparameter updates to noisy sample data. Specifically, RS makes it possible to improve estimation of the $\sigma^2_n$ noise hyperparameter in eq. (\ref{eq:gp_datalik}). This in turn can help improve estimation of the other hyperparameters in $\Theta$ early on in \BO{}. Recall that since $\sigma^2_n$ characterizes the external noise term $\epsilon_i$ in eqs. \ref{eq:noisy_var} and \ref{eq:noise}, this hyperparameter effectively accounts for errors that are independent of $\bm{x}_b$ in the GP surrogate model, i.e. errors in modeling $y_m$ that can arise due to an imperfect choice of kernel function or kernel hyperparameters. 

In theory there are at least two ways to incorporate RS experiments into a GP. The first is by incorporating the summary value of multiple experiments (i.e. the mean observed value from $n$ experiments). The second is to include all experiments individually. On the surface the first method seems attractive since the computational complexity of a GP is cubic with respect to the number of data points. In practice this approach does not yield desired results, instead the second method must be used. Put simply, this is due to the fact that a summary observation does not directly encode the observed variance. In contrast, using all data fundamentally alters $\fs$ and makes the hyperparameter learning process more effective. This point will be discussed further in the next section. 

In this work, acquisition functions for RS mirror those for the conventional SS approach, with the main difference being that multiple samples (instead of just one) are taken at the maximum of the acquisition function. Formal adaptations of MS acquisition functions could also be considered to constrain all samples to come from the same $\bm{x}_b)$ on a given optimization iteration.

\subsection{Comparison of MS and RS Effectiveness}
It is posited here that MS and RS can significantly improve the surrogate model's ability to capture underlying statistics (mean and covariance functions) of an underlying stochastic objective function, particularly early on in the \BO{} process. To gain some insight into the expected gains offered by these approaches, it is first necessary to define a notion of statistical consistency or `goodness of fit' for the surrogate model GP. Suppose a `ground truth' GP $\bar{\fs}$ exists to describe the stochastic objective function, which has some known kernel function $k$ and true hyperparameter values $\bar{\Theta}$. When performing \BO{}, another surrogate model $\fs$ with the same kernel function $k$ learns an estimate of the hyperparameters $\tilde{\Theta}$ from noisy data sampled from $\bar{\fs}$. Since $\fs$ and $\bar{\fs}$ use the same $k$, then $\fs$ is statistically consistent with $\bar{\fs}$ if $\tilde{\Theta} \rightarrow \bar{\Theta}$ as $n \rightarrow \infty$. Thus, the impact of using MS or RS can be understood (in this restricted regression scenario) by examining eq. \ref{eq:gp_datalik} following corresponding sample updates. 
		
\begin{figure}[!htbp]
            % \centering
            \captionsetup[subfigure]{farskip=-0.15cm,captionskip=-0.0cm,labelfont={large,bf}}
            \subfloat[]{%
                \includegraphics[width=0.85\linewidth]{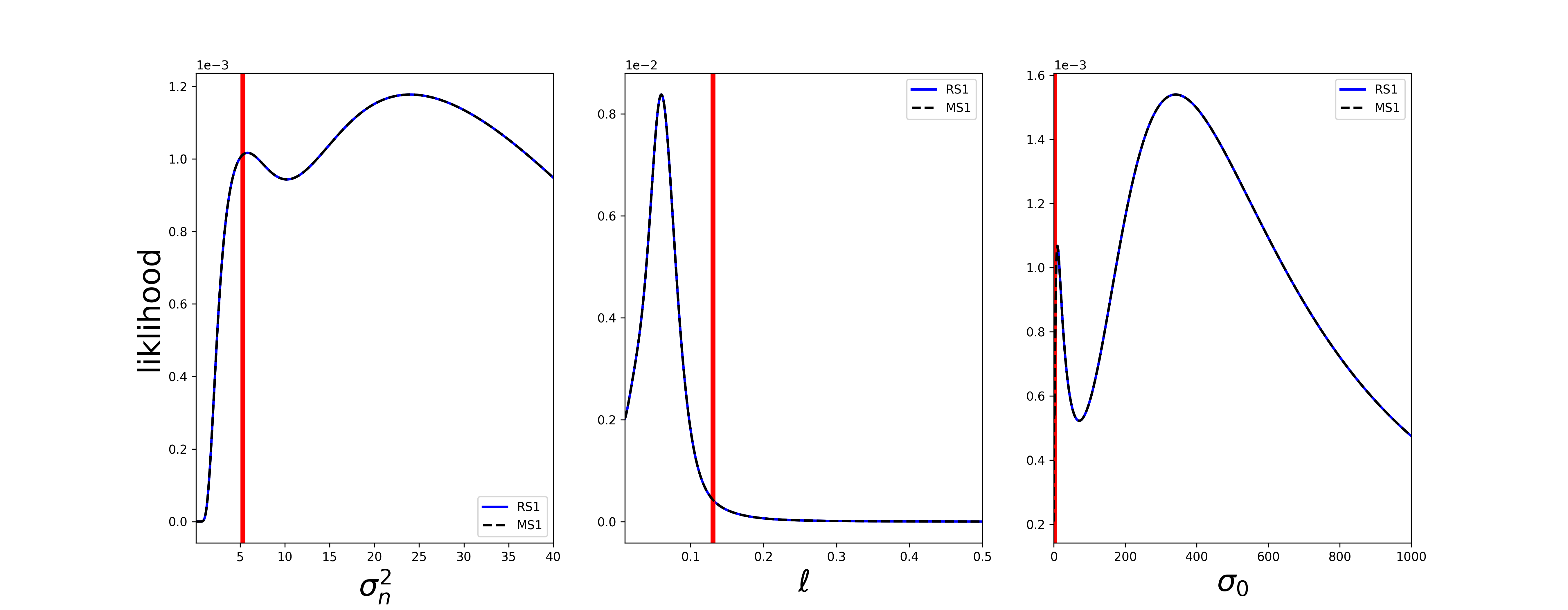}%
                \label{fig:RS1}%
            }\\
            \subfloat[]{%
                \includegraphics[width=0.85\linewidth]{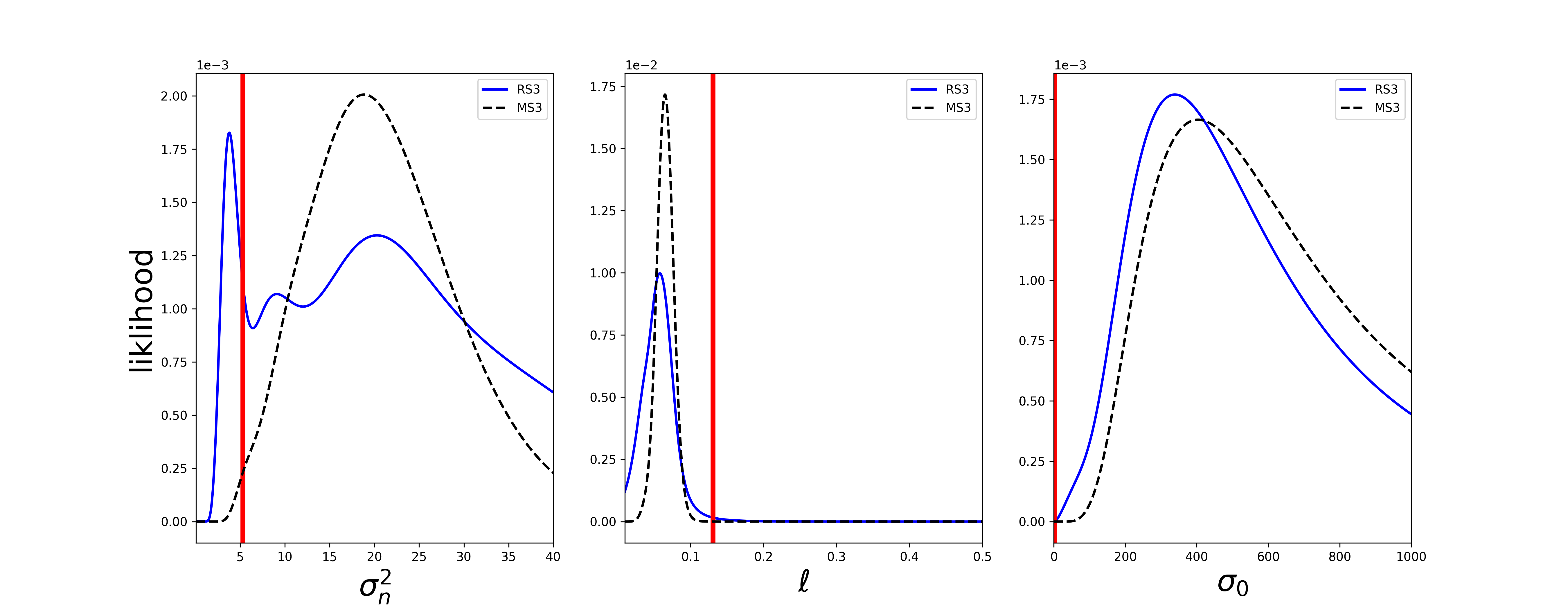}%
                \label{fig:RS3}%
            }\\
            \subfloat[]{%
                \includegraphics[width=0.85\linewidth]{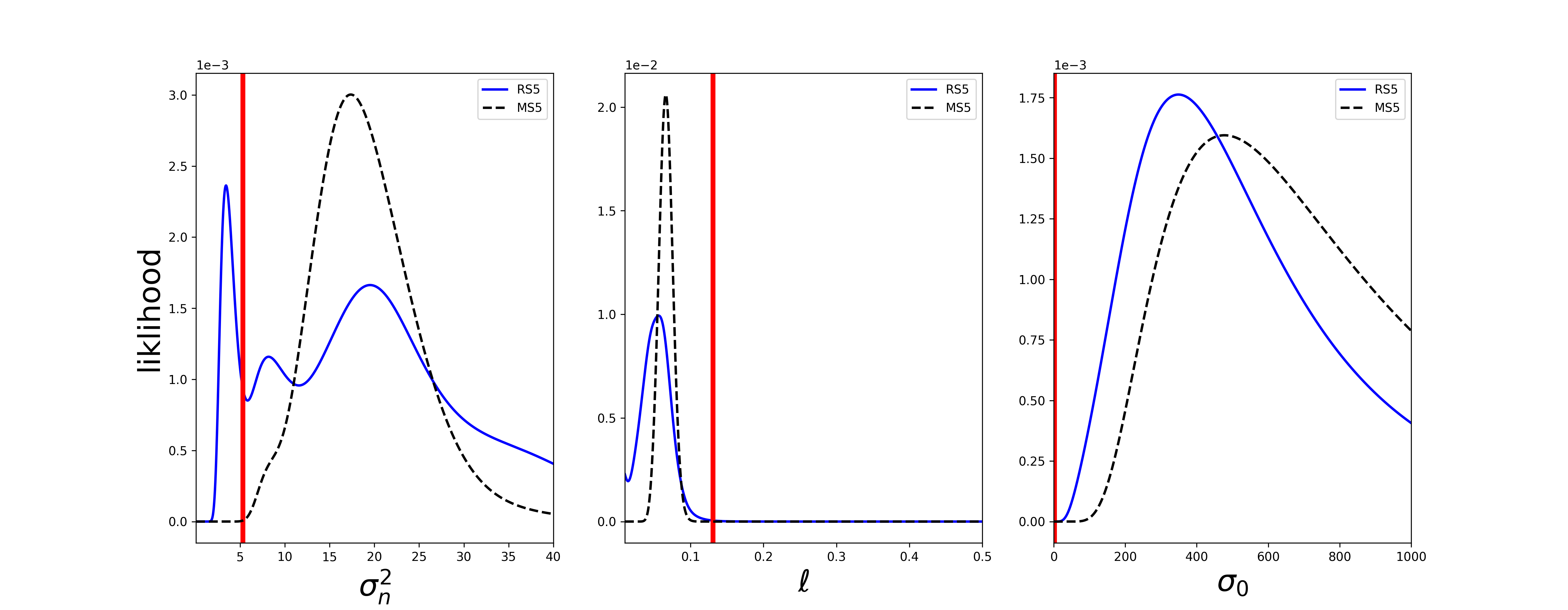}%
                \label{fig:RS5}%
            }%
            \caption{Numerical example of the effect of RS versus MS on hyperparameter learning with sparse data. The subplots represent: (a) $RS=MS=1$, (b) $RS=MS=3$, (c) $RS=MS=5$. Data was obtained using 100 simulations for each plot, the mean value is displayed. The vertical red lines indicate the true hyperparameter values, the solid blue line represents the RS curve, and the dashed black curve represents the MS curve.}
            \label{fig:rs_effect}
        \end{figure}

Fig. \ref{fig:rs_effect} shows results from a toy problem that examines this consistency idea for MS and RS search strategies, when only small data sets are available for initial seeding. A 1D `truth model' GP $\bar{\fs}$ with known kernel hyperparameters $\bar{\Theta}$ (red vertical lines) was used to generate noisy objective function evaluation data. Fifteen random data points were then used to construct an approximate `initial seed' GP surrogate model $\fs$ using the same kernel function, but with an estimated $\hat{\Theta}$. The same number of new sample points were then added to $\fs$ using either MS (with GP-UCB-PE) or RS (using UCB). The resulting GP data likelihoods from eq. (\ref{eq:gp_datalik}) were then evaluated following a single update over a mesh of hyperparameter values (varying one element of $\Theta$ at a time, while holding the other hyperparameters fixed to their initial values); the rows show results for RS=MS=1, RS=MS=3, and RS=MS=5, respectively, while the columns show variations of eq. (\ref{eq:gp_datalik}) with respect to each element of $\Theta$ for the Mat\'{e}rn kernel. The regression procedure was repeated 100 times with random initial seedings for each search approach, and likelihood curves are plotted for instances achieving median maximum likelihood values. 

The top row shows that both methods perform identically when RS=MS=1, as expected. The remaining results show that RS and MS both perform comparably on the $\ell$ and $\sigma_0$ parameters, and tend to remain biased early on in the \BO{} process when the other hyperparameters are fixed. For $\sigma_n^2$, MS has larger variance but a clear maximum far from the true $\sigma^2_n$ value, i.e. it is biased towards a larger $\sigma^2_n$ but has an appropriate amount of uncertainty to account for this. On the other hand, RS has a sharper local peak at the true $\sigma^2_n$ and more local optima, including a larger peak at a much larger $\sigma^2_n$ value. However, the local peak near the true value gets larger as the number of samples increases for RS, while MS reduces its bias more slowly with additional samples. This demonstrates the idea that, during hyperparameter optimization early on in the \BO{} process, RS has a better chance of identifying $\sigma_n^2$ value independently of the other hyperparameters, as new data points are sampled (i.e. as long as the best local optimum is found via MLE or MAP). It is also worth noting in this example that MS produces a more reliable and conservative estimate of the error uncertainty than RS, which tends to be more optimistic (as indicated the curvature of the likelihood functions). It is important to ensure that the hyperparameter error uncertainty is also correct, since the GP data likelihood is recursively updated by \BO{} in future sampling steps and thus must correctly steer the behavior of the surrogate and acquisition functions via hyperparameter updates. 

In practice, as alluded to in the previous section, this means that including a summary statistic is indistinguishable from a single experiment; while the mean observed value will be less responsive to outliers, it will not convey information about the variance of the data. In contrast using RS encodes information about the variance of the experiments into the kernel and fundamentally changes the likelihood of the hyperparameters.

A quick note on computational complexity is merited as it is a serious concern for GPs when data sets are large. There are methods that can make a GP `sparse' by distilling the data to fewer more important parts; this is good idea when the GP is static, but not when it is still being learned. In other words, if the GP is done being trained (i.e. the hyperparameters have been sufficiently learned) then the data can be made sparse to improve the speed of calculations. However, if the GP is still being learned, attempting to make the kernel sparse will modify the hyperparameter likelihood and will hinder the learning process. It is therefore recommended that when learning hyperparameters (especially when very little data is available, as in an expensive optimization problem like this one) the GP be left `dense', and only reduced after the hyperparameters are no longer being learned.

\subsection{Hybrid Repeat/Multi-point Sampling (HRMS)}
The preceding considerations suggest that some combination of RS and MS might yield the `best of both worlds' for \BO{}, which potentially enables more efficient and correct statistical modeling of noisy objective functions. Specifically, RS should be used to improve handling of sparse noisy samples early on in the process (i.e. to `anchor' the GP surrogate), while MS should be used to explore and exploit the GP surrogate more efficiently. This leads to a novel third sampling strategy, referred to here as \textit{Hybrid Repeat/Multi-point Sampling (HRMS)}. In this approach, the MS number allows \BO{} to vary the total number of locations to investigate at each iteration, while the RS number allows \BO{} to vary how many times to sample each location at each iteration. The remainder of the paper uses the convention that MS=$m$ means $m$ locations in $\bm{x}_b$ space will be simulated in parallel, and RS=$l$ means that the same location in $\bm{x}_b$ space will be sampled $l$ times. 

In general, both $m$ and $l$ could vary over the course of the \BO{} process, although compatible acquisition functions should be used for both the MS and RS portions. In this work, it is assumed that $m$ and $l$ are selected and fixed a priori for HRMS \BO{}. For $m>1$ and any $l\geq 1$, an MS acquisition function is used to select $m$ distinct sampling locations, at which $l$ repeated samples are evaluated. Due to the highly application-dependent nature of the problem, it remains a challenging and open research problem to determine how exactly $m$ and $l$ should be varied to achieve optimal \BO{} performance. For instance, these parameters could be selected to minimize time to find $\bm{x}^*_b$ in the fewest iterations possible. Alternatively, they could be selected to minimize the total number of objective function evaluations. The experimental results presented in Sec. \ref{sec:results} will provide some basic insights into this question in the context of the aerial combat simulation application.

    \section{Simulation Results}\label{sec:results}
    This section presents the results of simulation studies that were conducted to investigate the performance of the different \BO{} acquisition functions and sampling strategies on two different test problems. 
    First, evaluation results for a standard benchmark optimization problem are presented, primarily to validate the newly proposed RS and HRMS sampling techniques on a non-trivial test case with a known solution. 
    Then, results for the simulated air combat decision-making AI optimization problem described in Section \ref{sec:methodology} are presented, along with an analysis and discussion of the outcomes of \BO{} on the AI pilot parameter adaptation process.

Three different acquisition functions were evaluated in each test case using SS, MS, RS and HRMS sampling techniques: for SS and RS, these were Expected Improvement (EI)~\cite{Jones1998}, upper confidence bound (GP-UCB)~\cite{Srinivas2010}, and Thompson Sampling (TS)~\cite{Thompson1933}; for MS and HRMS, these were the corresponding batch sampling forms: q-EI~\cite{Wang2016}, GP-UCB-PE ~\cite{Contal2013}, and multi-draw TS. The different levels of RS and MS used are RS$=\{1,3,5,10\}$ and MS$=\{1,3,5\}$ (where SS corresponds to RS=MS=1 and HRMS corresponds to RS$>1$ and MS$>1$). 
The GPML toolbox~\cite{Rasmussen2006b} is used for GP representation and hyperparameter inference. 

\subsection{Ackley 3D Function Experiments}
\begin{figure}[!htbp]
          \centering
          \includegraphics[width=1.0\textwidth]{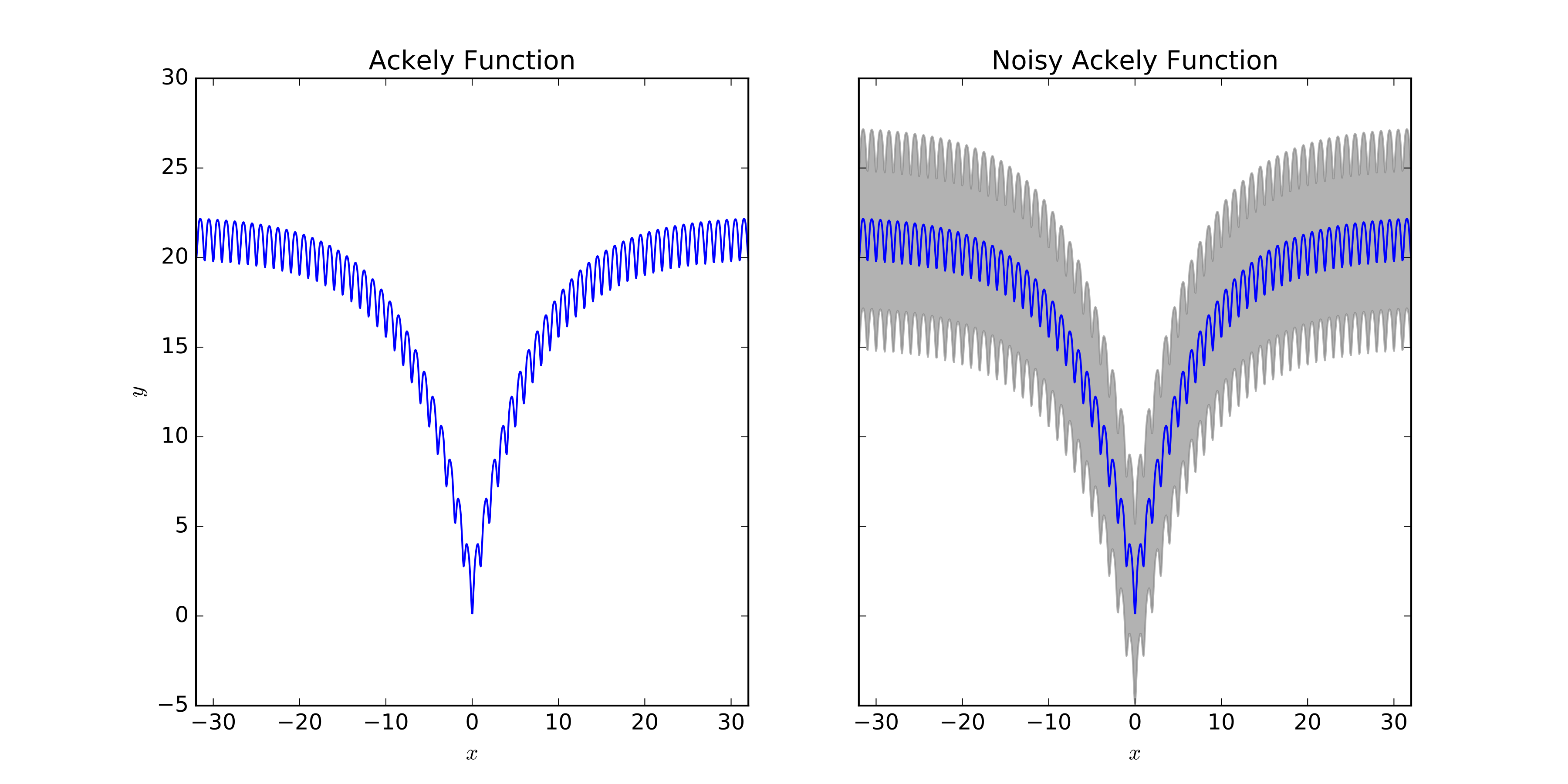}
          \caption{The left plot is the standard Ackley function. The plot on the right is the standard Ackley function with noise variance $\sigma_y^2=25$}
          \label{fig:ackley}
    \end{figure}

To validate the usefulness of RS and HRMS, it is useful to apply these sampling methods to a standard benchmark optimization problem with a known minimum. The benchmark used here is a noisy version of the Ackley function~\cite{Back1996-jp},which is defined in $n$ dimensions and has its largest local minimum value of $0$ at $\bm{x}=\bm{0}$, where $\bm{x}$ is an $n$-dimensional input vector. Figure~\ref{fig:ackley} shows the standard and noisy versions of the Ackley function for $n=1$. The deterministic base Ackley function has multiple periodic local minima that are closely spaced, and so this function is highly non-trivial to optimize using standard numerical optimization techniques. 
The noisy version of the Ackley function thus serves as an even more interesting and challenging benchmark for assessing \BO{} techniques. 

While \BO{} techniques are theoretically able to cope with the presence of multiple local expected minima,  
a poorly tuned or selected \BO{} algorithm (like any other optimization technique) can easily get stuck at a local minimum and fail to explore the solution space. 
These issues become particularly challenging in high-dimensional settings, where sensitivity increases to initial seed conditions and poor local minima in the GP log-likelihood or log-posterior. 
Hence, this benchmark problem can provide some insight into how well different \BO{} strategies can explore complex stochastic objective functions, as well as accurately and repeatably represent them through the surrogate function.
    
The noisy Ackley function was optimized on 10 independent runs for each RS/MS \BO{} configuration with $n=3$ and $\sigma_y^2=25$, using 20 random initial seed points in the ranges $x_1=(-32.768, 32.768)$ $x_2= (-12.21, 32.768)$, and $x_3=(-32.768, 5.14)$ for each dimension of the search space (thus bounding the largest local minimum for this problem at $\bm{x}=\bm{0}$). MAP hyperparameter estimation was used with the Laplace approximation, with Mat\'ern $3/2$ kernel hyperpriors set to uniform distributions between $e^{-1}$ to $e^2$. 

    \begin{figure}[!htbp]
        \centering
        \includegraphics[width=1.0\textwidth]{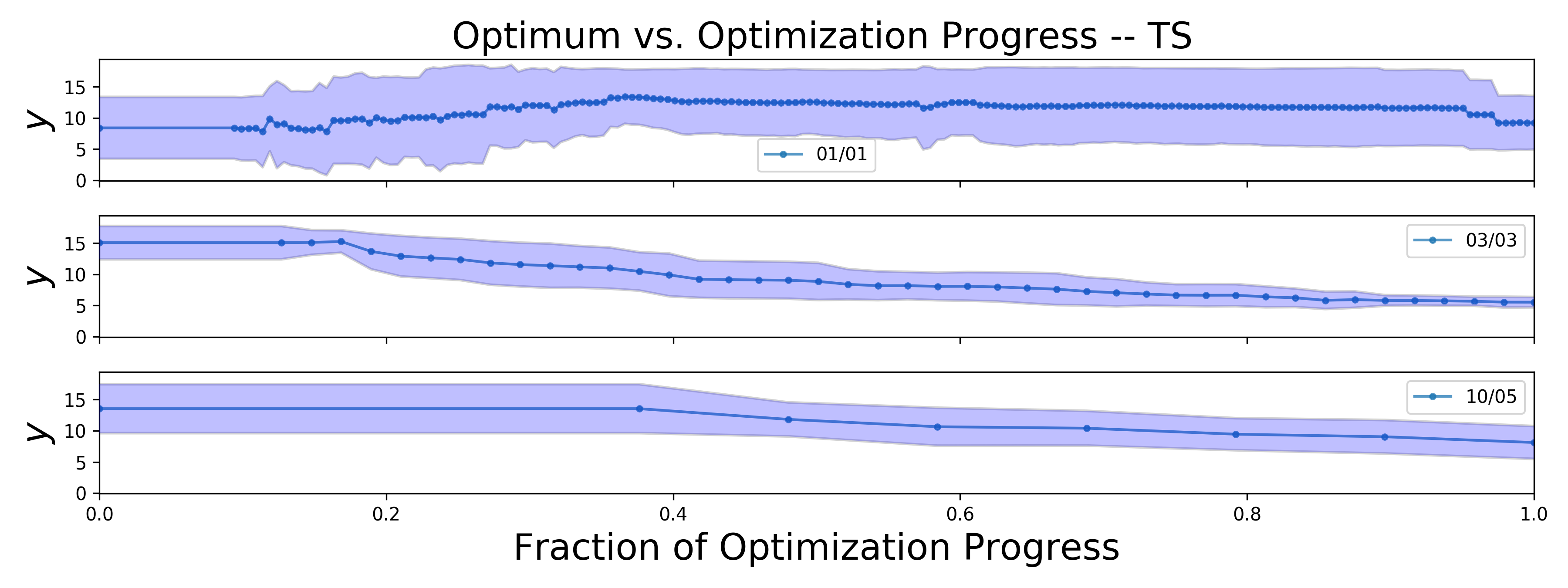}
        \caption{Plot illustrating the convergence of three different RS/MS sampling strategies (i.e. the top corresponds to RS=1, MS=1 strategy) in optimizing the Ackley function. Shown are the results of 10 experiments for each of the three RS/MS configurations. The solid blue line represents the mean objective value obtained during optimization, and the shaded area is one standard deviation.}
        \label{fig:convergence}
    \end{figure}

Figure \ref{fig:convergence} is a convergence plot that illustrates the effect of HRMS on this problem where three different HRMS sampling strategies are applied using a time limit of 1 hour (the number of function evaluations and iterations are not equal due to the complexity of different strategies). The main result of interest is found by comparing the right-hand side of each sub-plot. For example the 03/03 approach yielded a better minimum than that of the 01/01 approach, it also had less variance between independent optimization attempts. This illustrates that there is utility in using HRMS on a function with high noise. This finding becomes even more critical when considering that in true black-box optimization there is no way to know when the algorithm converges, and thus using an HRMS approach (instead of SS) yields a more reliable and more accurate solution.

It should be noted, however, that in order to make a \emph{fair} comparison between the different \BO{} methods, the inherently different baseline computational costs of each approach need to be taken into account. Whereas, optimization methods are typically compared by number of iterations until convergence, this comparison does not translate to the current comparative setup for \BO{}, since some approaches might use many more function evaluations per iteration (i.e. RS=10, MS=5 will use 50 function evaluations per optimization iteration, while RS=1, MS=1 will only use 1). Instead, the different \BO{} approaches are more appropriately compared based on a fixed total number of function evaluations; also, since the real interest is the final result of the optimization further figures will reflect the final results and not show typical convergence plots. 

    \begin{figure}[!htbp]
          \centering
          \includegraphics[width=1.0\textwidth]{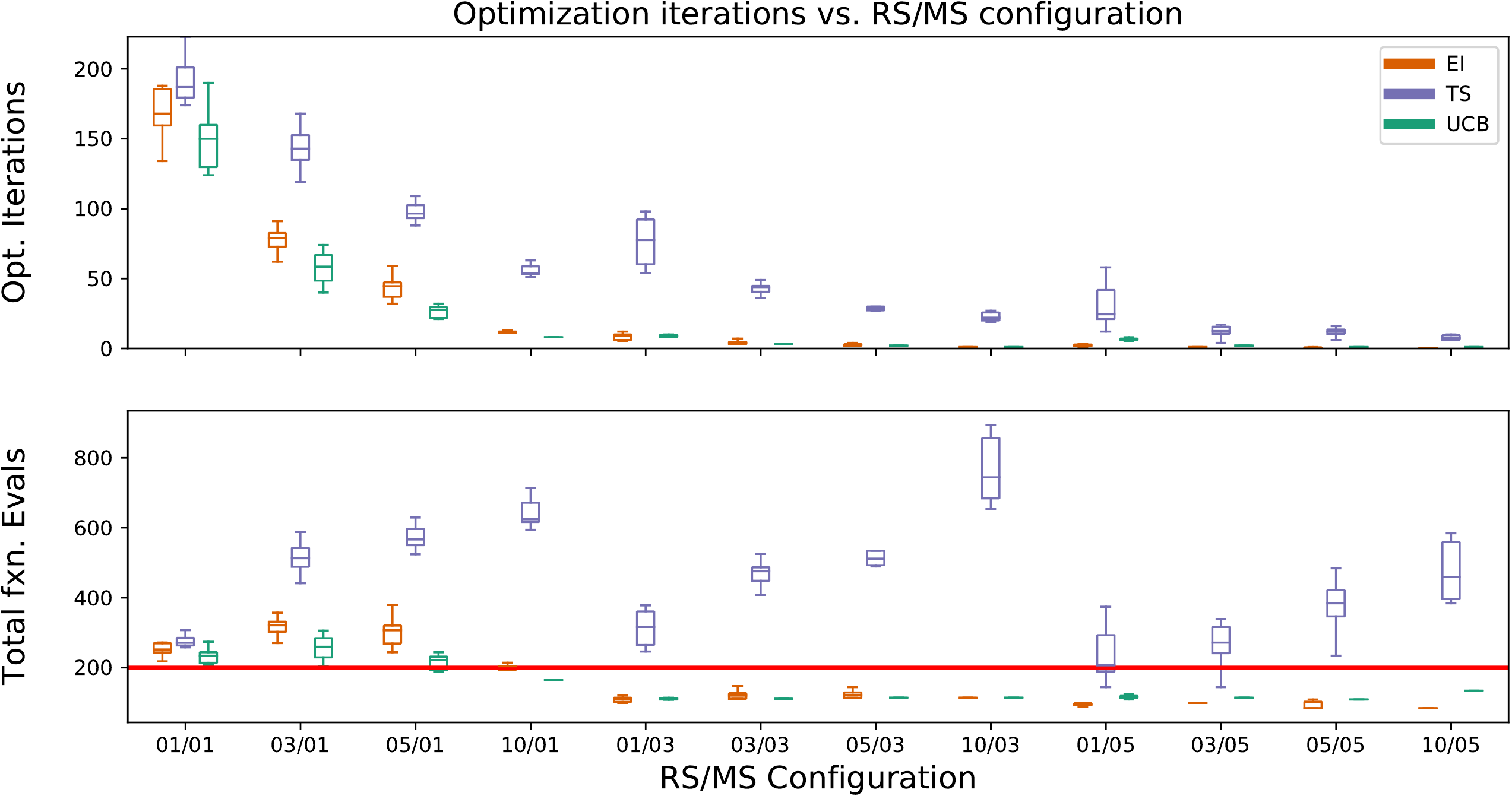}
          \caption{Plots of the number of optimization iterations and total function evaluations for each acquisition function and HRMS configuration. These numbers were obtained by using optimizing the Ackley function for a fixed time budget of 1 hour. The red line shows the point at, or below, which the parameter estimates and optimum were taken to make Figure~\ref{fig:3d_eval_comparison}. From left to right the boxes represent EI, TS, and UCB.}
          \label{fig:3dexp_dist_its}
    \end{figure}

\begin{figure}[!htbp]
          \centering
          \includegraphics[width=1.0\textwidth]{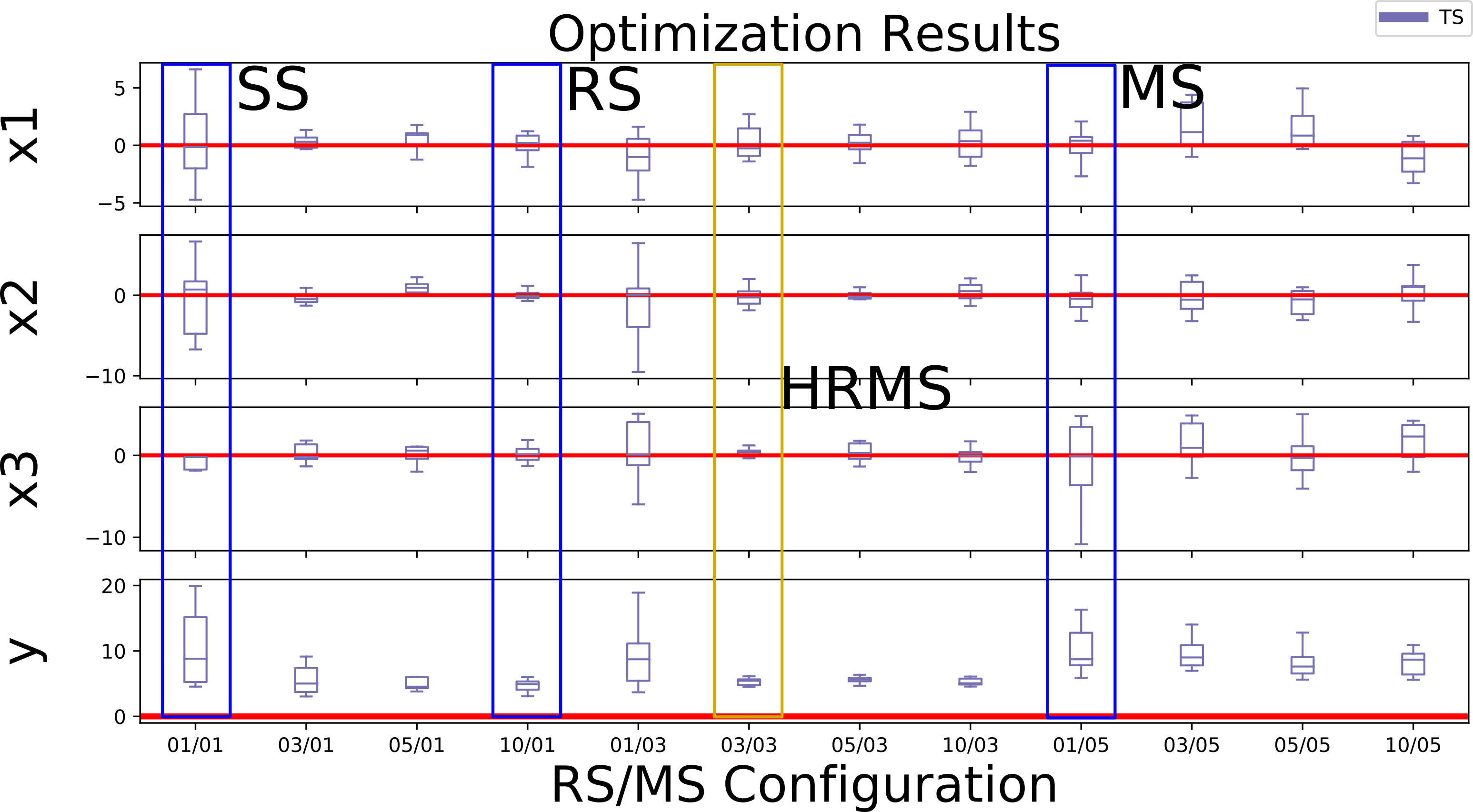}
          \caption{Box plots of the $x_1,x_2,x_3$ and $y$ values of estimates of the Ackley optimum found by TS for each HRMS configuration. The $x$ axis represents different RS/MS configurations. The data points reflect the final parameter estimate of an optimization run. The red horizontal line denotes the truth value. All estimates were obtained after 200 (or fewer) total function evaluations.}
          \label{fig:3d_eval_comparison}
    \end{figure}

Figure~\ref{fig:3dexp_dist_its} illustrates the difference between optimization iterations and total function evaluations across RS/MS configurations and acquisition functions. These results were generated after running optimization for 1 hour, in order to highlight the different speeds at which each \BO{} method runs. Generally, SS was the fastest approach yielding more function evaluations. Due to the expense of calculating the MS locations, the total number iterations for MS decreased as more evaluations were added per iteration. Notice that UCB had much fewer evaluations as MS increased, due to the high expense of the GP-UCB-PE algorithm. On the other hand, the cost of drawing MS random functions when using TS was relatively inexpensive (given that the number of sampled points is `small enough'). Note that the red-line on the bottom plot shows the `cut-off' point at, or below, which the estimates were taken in order to make Figure~\ref{fig:3d_eval_comparison}, so that the different methods could be fairly compared for a given number of function evaluations.
    
\begin{figure}
        \centering
        \includegraphics[width=1.0\textwidth]{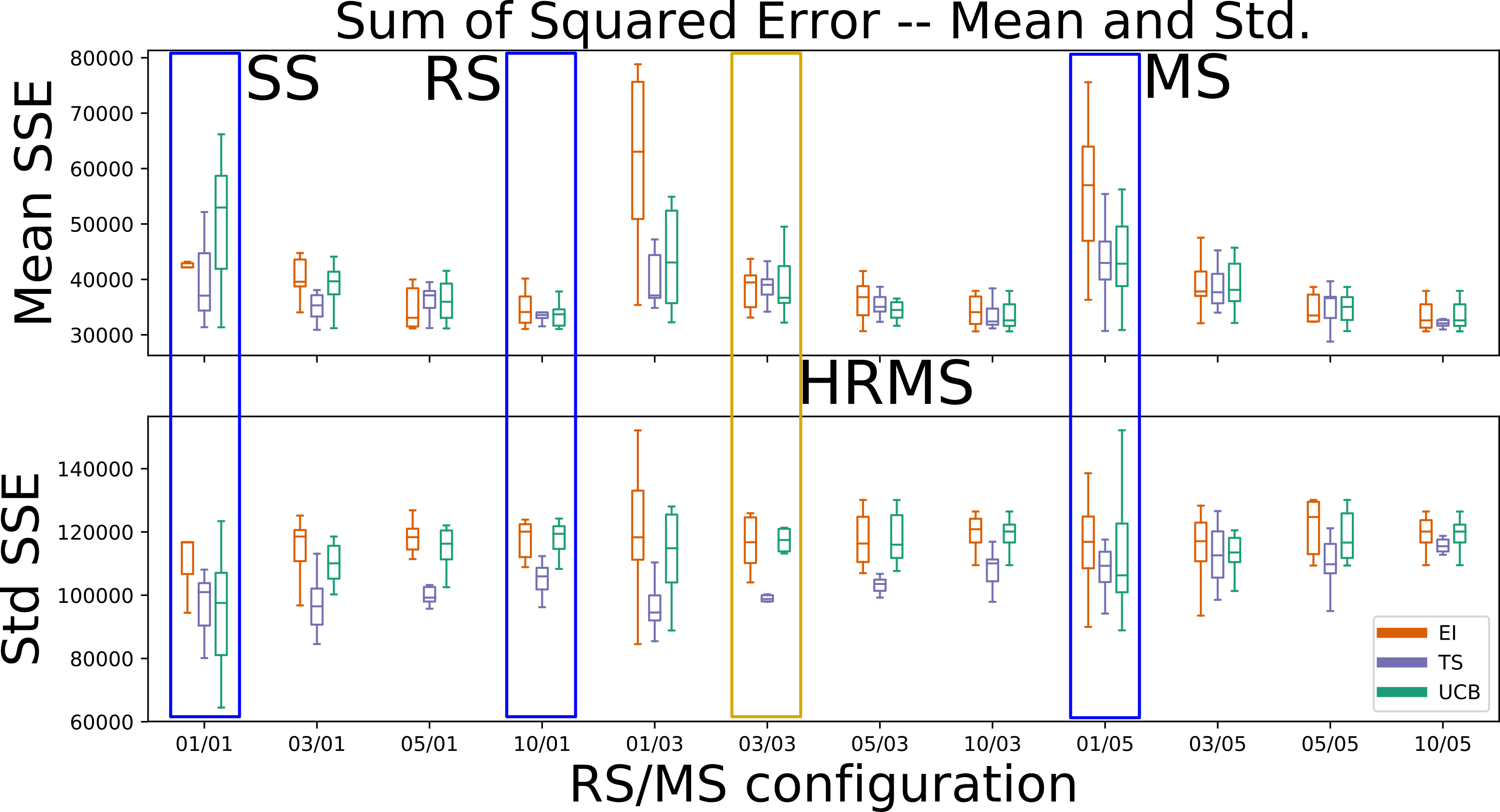}
        \caption{Reduction in the sum of squared error in the optimization of the Ackley function for different HRMS configurations: (top row) SSE of the GP surrogate mean predictions compared to the ground truth; (bottom row) SSE of GP surrogate $\sigma$ compared to the ground truth. Blue boxes highlight the standard SS and MS implementations; orange box highlights an example of \emph{one} HRMS configuration.}
        \label{fig:3dSSE}
    \end{figure}

The results of the optimization trials are summarized in Figures~\ref{fig:3d_eval_comparison} and~\ref{fig:3dSSE}, in terms of the outcomes of the estimates for each of the inputs and the optimum value for a fixed set of 200 function evaluations per method (or fewer if more could not be finished within the one-hour limit). Note that only the results for the Thompson sampling acquisition function are shown here; the results for the other acquisition functions show similar trends, but are not displayed here. 

The areas highlighted by blue boxes show the standard approaches using SS and MS. The orange box represents one example of the outcome of optimization using HRMS. Notice how the RS and HRMS configurations, in general, exhibit smaller variance and closer proximity to the true value. From this figure we can conclude that there are RS/MS configurations that yield more reliable and accurate optimization results than those of SS or pure MS.

In Figure \ref{fig:3dSSE}, the sum of the squared errors (SSE, also called the residual sum of squares) between the final surrogate GP model and a `ground truth' objective model is shown. The SSE is calculated using
\begin{align}
    \text{SSE} = \sum_{i=1}^{n}{(v_{i,predicted}-v_{i,true})^2}
\end{align}

\noindent
where $v$ is the value of interest at location $i$ (i.e. the mean or variance).  In other words the SSE is the sum over $n$ locations of the squared error between the predicted and true values. Here, the `ground truth' model is another higher fidelity GP fit with 10,000 training samples drawn from the noisy 3D Ackley function; this allows for consistent comparison of the GP surrogate model mean and variance functions for the full range of input parameters. 

A smaller SSE denotes a closer fit to the true objective function; an SSE with smaller variance reflects better repeatability.  Notice the universal improvement of the RS and HRMS methods over the standard SS and MS configurations in modeling the objective function mean. The SSE of the standard deviation reduces variance with the HRMS method, but does not seem to get closer to that of the true function. Although this behavior is not yet fully understood one hypothesis is that this may be due to the sparsity of the experimentation using \BO{}, as well as the fact that there is higher uncertainty in areas that haven't been explored. In other words the mean representation is better but the variance is still large in unexplored areas, similar to what is seen in iteration 4 of Figure \ref{fig:UCB_example}.

To summarize some of the main insights from these results:    
    \begin{enumerate}
        \item While it is \emph{possible} to obtain reasonable optimization results using the standard SS and MS strategies, the solutions are highly sensitive and have high variance (as seen in figures \ref{fig:convergence} and \ref{fig:3d_eval_comparison};
        \item Using HRMS, it is possible to not only outperform the accuracy of the predictions from SS and MS sampling approaches, but to obtain a better representation of the true objective function as well (as seen in figure \ref{fig:3dSSE}).
        \item HRMS often realizes these improvements with \emph{fewer} total function evaluations than standard sampling approaches (see figure \ref{fig:3dexp_dist_its}). This is a characteristic that is critical in optimization of `expensive' objective functions (such as high-fidelity air combat simulations).
    \end{enumerate}
Having demonstrated the efficacy of RS and more generally HRMS on a highly noisy objective function, we next assess whether these insights translate to the adaptive training of AI pilots for the simulated air combat application.

\subsection{Air Combat Simulation Experiments}
The AI pilot here has many parameters that can be modified (see Table~\ref{tab:agent_params}). However, only 2 parameters in particular were found to have significant effects on the outcome of the combat simulations for the one-on-one engagement scenario described in Sec. \ref{sec:methodology}: $launch$ (launch delay) and $intspeed$ (intercept speed). Hence, the simulation studies only focus on optimizing the two parameters $\bm{x}_b = [launch, \ intspeed]^T$ for the blue AI pilot to minimize the $TTK$ metric, while fixing the others at their default values. 

The simulation was optimized during 10 independent runs for each RS/MS \BO{} configuration using 20 random initial seed points in the ranges $x_1=(0,5)$, and $x_2= (0,500)$ for each dimension of the search space. MAP hyperparameter estimation was used with the Laplace approximation, and the Mat\'ern $3/2$ kernel hyperpriors set as follows: the mean as $\mathcal{N}(0,100^2)$, and the covariance parameters as $\mathcal{U}(\log 1, \log 3)$. 
    
    \begin{figure}[!htbp]
          \centering
          \includegraphics[width=1.0\textwidth]{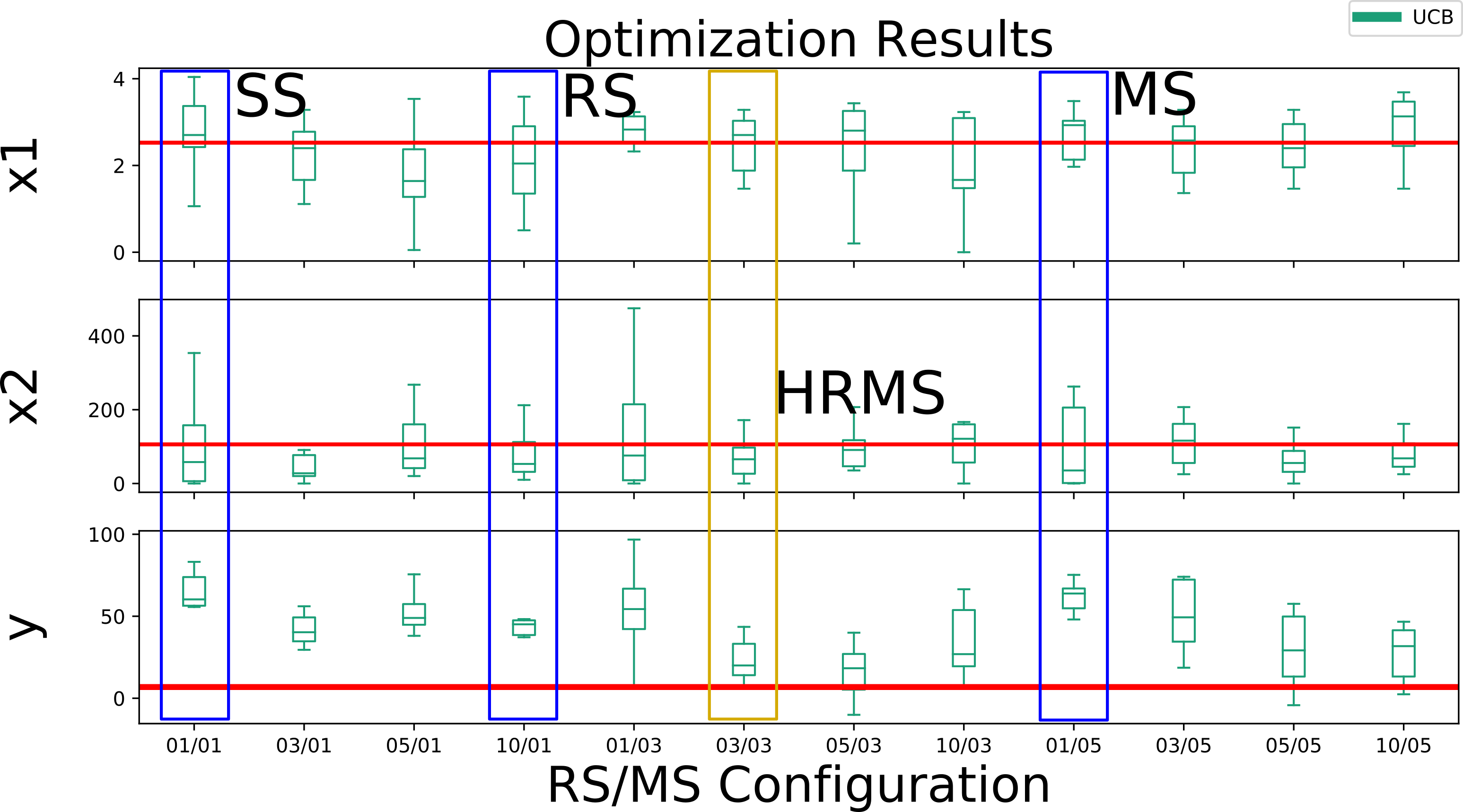}
          \caption{Box plots of the $x_1,x_2$ and $y$ values during optimization of TTK for each acquisition function and HRMS configuration. The red horizontal line is the ground truth value. These results are compiled from the results of 10 separate optimization runs at the corresponding HRMS configuration.}
          \label{fig:2d_time_plot_comparison}
    \end{figure}

Figure~\ref{fig:2d_time_plot_comparison} shows the estimated locations of the optimum $TTK$, as well as the `ground truth' optimum location estimated from a brute force grid search over the blue parameter space $\bm{x}_b$ using 10,000 simulations (results for the for the UCB acquisition function are shown only; the remaining acquisition functions show similar trends and are omitted to reduce clutter).
The estimates for $\bm{x}_b$ and $y$ grow tighter together, and closer to the ground truth, as both the RS and MS configuration numbers become greater than 1. The configurations marked by colored rectangles highlight that methods using solely SS, RS, and MS (shown in blue), underperform the method that combines both RS and MS greater than one (yellow). This finding is similar for the EI and TS functions as well, and reflects the results found for the 3D Ackley function. From this figure we can conclude that there are HRMS configurations that yield more repeatable optimization results than SS, and that neither RS or MS alone is clearly better.

    \begin{figure}[!htbp]
          \centering
          \includegraphics[width=1.0\textwidth]{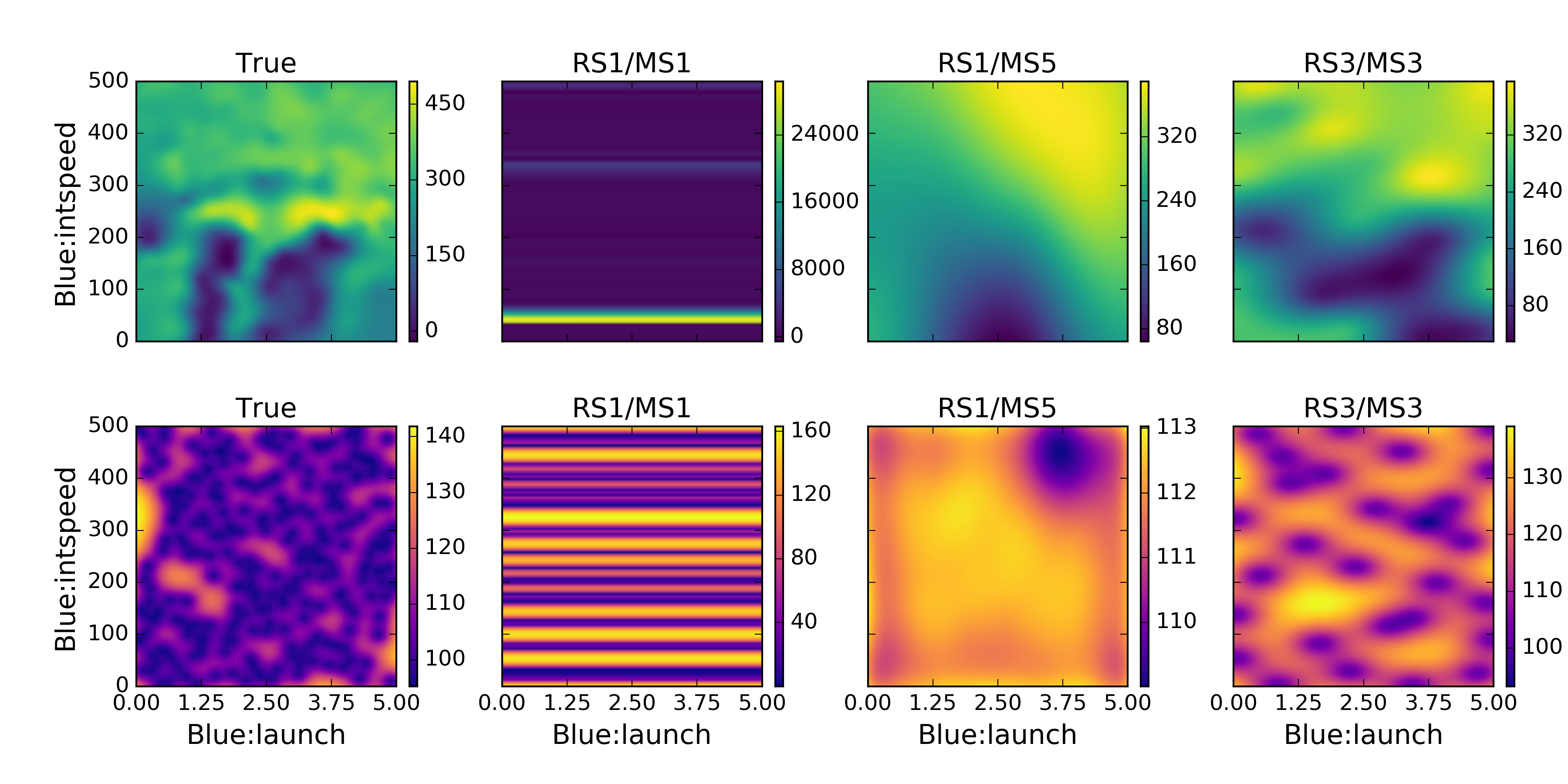}
          \caption{Table of figures illustrating the effect of combined RS/MS sampling using the UCB acquisition function to optimize TTK. Top row is $\mu$ bottom row is $\sigma$ for the GP surrogate model. The leftmost column is the truth surface obtained by high density sampling and fitting a GP to the data. The following columns show some example results from optimization runs using the indicated values for RS and MS. Each of the final 3 columns represents the optimization solution after 1 hour}
          \label{fig:sample_instability}
    \end{figure}

\begin{figure}[!htbp]
        \centering
        \includegraphics[width=1.0\textwidth]{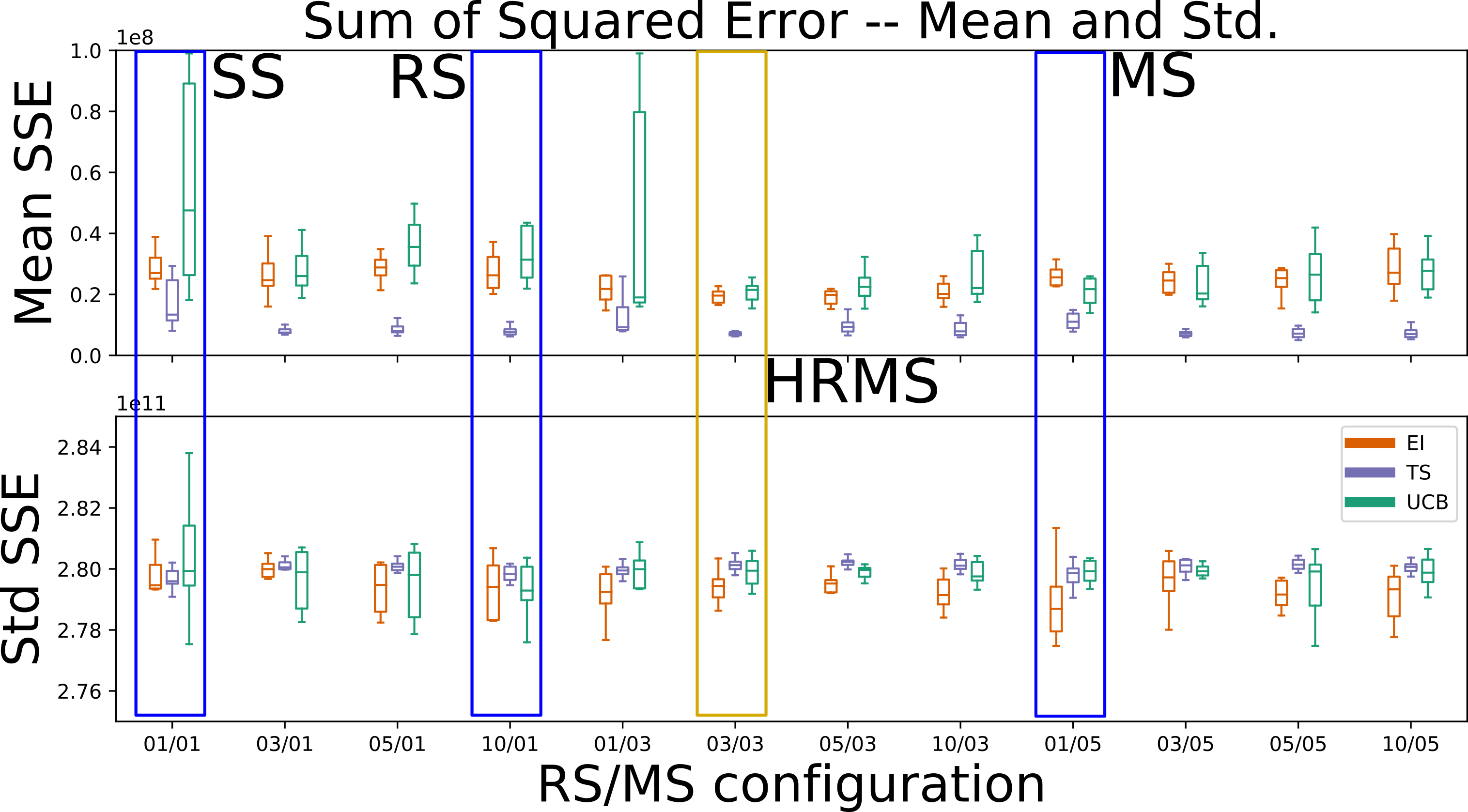}
        \caption{Plot illustrating the reduction in the sum of squared error in optimizing TTK for different HRMS configurations. The top figure is the SSE of the mean predictions compared to the ground truth. The bottom figures is the SSE of $\sigma$ compared to the ground truth.}
        \label{fig:2dSSE}
    \end{figure}

Figure~\ref{fig:sample_instability} depicts some examples of the final GPs obtained during time limited optimization (again, allowing faster methods to use more evaluations/iterations) for three different HRMS configurations. The far left column is the `ground truth GP' model that is obtained by training with several thousands of samples over the input space. The key insight is that the RS3/MS3 strategy yielded a GP that better represents the ground truth. This confirms some of the observations from the previous study of the Ackley function. Also as in the previous study, Figure~\ref{fig:2dSSE} illustrates that using HRMS improves the surrogate model as well as the optimization outcomes. These two phenomena are linked: a better surrogate model leads to more accurate and reliable optimization results. Returning to the high level goal of this research: training an AI with behavioral parameters to optimize a combat objective, and to provide tools for being adaptive. Figure~\ref{fig:2d_time_plot_comparison} demonstrates that \BO{} is useful in this application. It shows that using HRMS results in more repeatable identification of the optimum AI behaviors in the simulated air combat.
	
Figure~\ref{fig:sample_instability} shows that the combined RS3/MS3 strategy (column 4) yields a more accurate surrogate representation of the true objective function. This surrogate model gives insights about how the expected value of the TTK changes over the entire behavioral space for the blue AI agent. There seem to be two large areas with lower $intspeed$ and moderate $launch$ that provide the best performance for blue. Conversely, it appears that the higher $intspeed$ configurations do not yield very good results. Looking at the full playback of recorded telemetry and decision data at the sampled simulation points in/near these regions helps gather more insight as to why. 

    \begin{figure}[!htbp]
        \centering
        \includegraphics[width=0.90\textwidth]{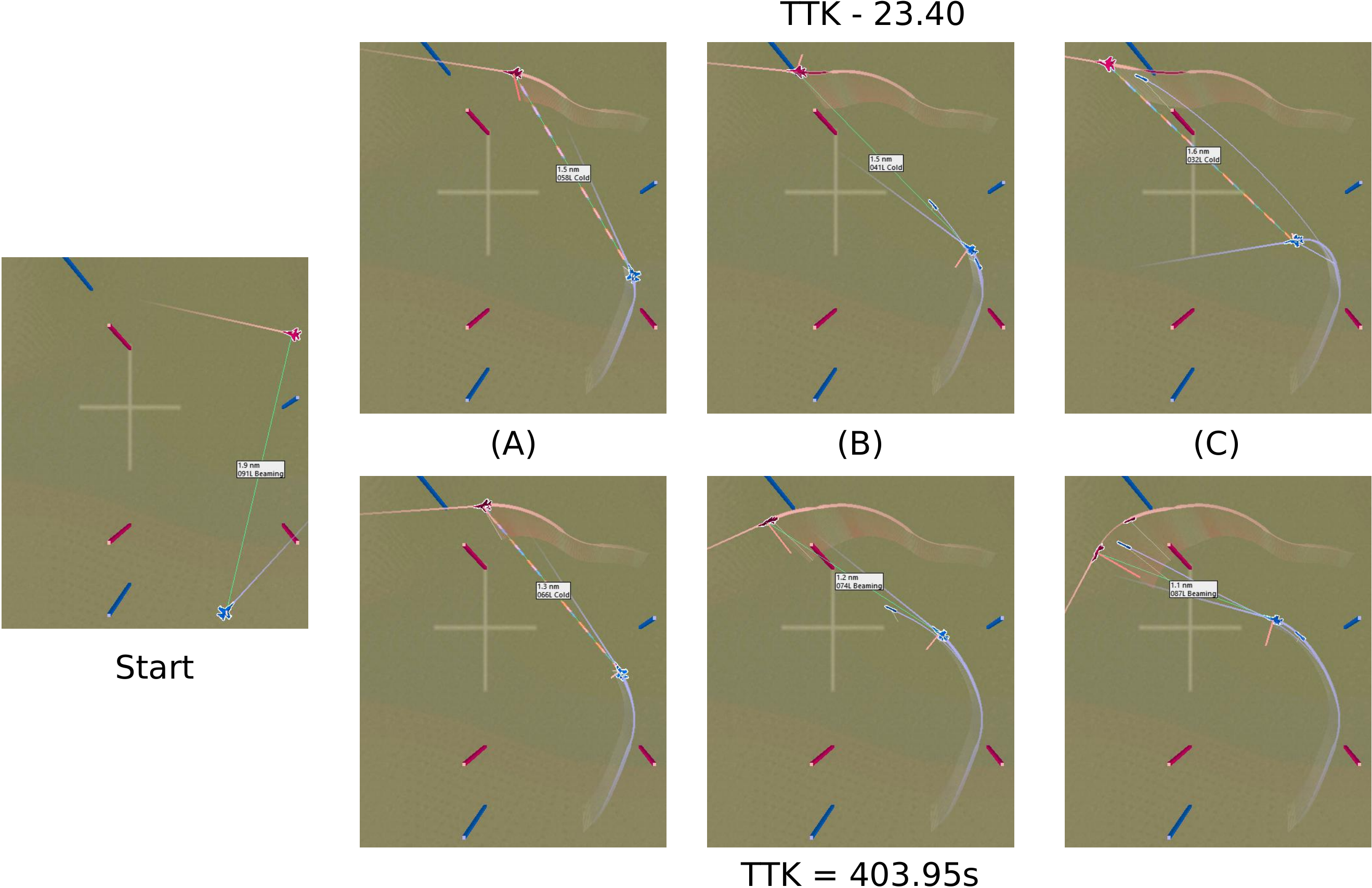}
        \caption{Comparison between $\bm{x}_b=\{2.60,125.0\}$ with associated $y=TTK=23.4$ secs (top) and $\bm{x}_b=\{3.00,450.0\}$ with $y=TTK=403.95$ secs (bottom). The far left image shows the simulation start which is identical for both scenarios. }
        \label{fig:tacview}
    \end{figure}  

Figure \ref{fig:tacview} shows the outcomes of two different simulations (rendered here in the open source TacView air combat simulation visualizer): a scenario with lower $launch$ and $intspeed$ values that has good results; and another scenario with higher $intspeed$ that yields poor results because the blue agent cannot regain lock after launching a mid-range missile. In this simulation the aircraft must have an active lock on the target to continue guiding the missile.
The images in the (A) column show the point at which the blue agent locks on and launches its mid-ranged missile. 
The images in column (B) show that in both scenarios blue loses lock on red. However, in column (C) the blue with lower intercept speed is able to regain lock;  on the bottom blue is not able to maneuver in order to regain lock in time, and without lock the missile is not longer guided to it's target. As a result of blue missing red in the bottom scenario, it must spend a significant amount of time maneuvering again to regain lock on red. By lowering $launch$ and $intspeed$, blue is able to maintain a lock on red and successfully guide its missile much more consistently on the first pass.  

While the RS3/MS3 strategy allowed some insight into the combat outcomes, the surrogate functions identified with SS and to a lesser extent RS1/MS5 (columns 2 and 3) offer less reliable insight into the true $TTK$ function. The improved capability of predicting what might happen in untested conditions allow the AI to identify the expected optimum behavior, but also have more capacity to intelligently adapt by being able to predict outcomes at any other location in the behavioral space. Another ramification is that these models are, to some extent, interpretable by humans. It is easily observed from the 2d plots in this case, but might also be interpretable in higher dimensions by utilizing more advanced visualization techniques.

\subsection{Discussion}
The simulations results show that \BO{} with HRMS reliably finds optimum behaviors for the adaptive AI pilot, even given the inherent volatility in aerial combat. Compared to the pure RS and pure MS sampling conditions, HRMS-based \BO{} is generally able to learn a more accurate model of the true objective function with fewer total combat simulations using any of the acquisition functions considered. The resulting model thus provides good predictions of engagement performance over the possible behavior/AI parameter solution space, thus allowing the AI to efficiently adapt behavior. 

These results have several implications for minimizing the cost and improving the efficiency of mixed human-AI team LVC training. 
As already shown for AI-vs-AI training, parallelized simulation runs can be easily exploited to accelerate adaptation and behavior exploration via RS, MS or HRMS. 
Although not explicitly demonstrated here, recent work in the computational cognitive science literature suggests that \BO{} techniques are also quite effective tools for automating intensive training programs focused on evaluating and improving human performance in skills-based tasks \cite{Khajah-CHI-2016}. 
As such, \BO{} methods can be readily adapted to enhance human-vs-AI training as well. 
For instance, the blue force AI in the air combat mission simulation can be replaced by a human pilot, so that the adaptive AI now is responsible for the red agent instead. The goal of the \BO{} learning engine in this case is then to maximize its performance (TTK or other metrics) against the human pilot with as few simulations as possible. The GP surrogate model produced by the learning engine would thus also serve as a valuable tool for predicting the human pilot's performance against various red AI configurations. The learning engine could also be augmented to incorporate variations in combat environment parameters, so that the $\bm{x}$ search space for \BO{} includes factors such as weather, wind, terrain, visibility, and others in addition to red agent parameters. 
For single human pilot training, the RS, MS and HRMS methods can be used to repeatedly generate block test conditions, whose elements are then evaluated sequentially with the same pilot (possibly in randomized order). 
To fully leverage the parallelizability of \BO{} sampling and exploration, these sampling techniques could also be used to generate an `average human' performance model using the outcomes of parallel engagements with multiple human pilots (e.g. who simultaneously engage the AI one-on-one in different instances of the same mission). 
It is interesting to note that the stochastic \BO{} surrogate model can theoretically accomodate the high performance variability likely to be encountered among human pilots (due to individual differences, learning effects, etcetera). An interesting open technical question in this case is the extent to which \BO{} needs to be modified (if at all) to cope with possible non-stationarities in the objective function (i.e. changing performance statistics with respect to time and/or $\bm{x}$ parameters). 

    \section{Conclusion}
This paper studied and further developed \BO{} techniques for automatically tuning the decision-making parameters of an AI-controlled dog-fighting agent, so that it could learn to optimize performance against an intelligent adversary. Performance was measured by a non-closed form stochastic objective function evaluated on full-length simulated combat engagements and was modeled over the full agent behavior parameter space by a Gaussian Process (GP) surrogate function. As per the standard \BO{} formualtion, the GP surrogate function was used to investigate new simulations at different parameter settings that most likely yield optimum performance. However, due to the highly volatile nature of the underlying objective function in this application, standard \BO{} methods tend not to construct accurate surrogate models during optimization, and thus tend not to yield reliable optimization results. Novel repeat sampling (RS) and hybrid repeat/multi-point sampling (HRMS) methods for \BO{} were developed to address these issues. Numerical simulations on a benchmark optimization problem and a one-on-one AI-vs.-AI dog-fighting application scenario showed that \BO{} with HRMS generally led to more accurate surrogate models and more efficient adaptation to the optimum agent parameters. 

\BO{} provides an attractive solution for performing optimization on expensive `black-box' objective functions, due to its ability to make predictions about the value of the objective function in unexplored locations. In many respects, the adaptive nature of \BO{} closely resembles the `explore-exploit' behavior of reinforcement learning \cite{Russell-AIBook-1995} -- with one major difference being that \BO{} is applied here \emph{in between} full engagements to optimize a holistic cost function in an `offline' sense, whereas reinforcement learning strategies might typically be applied \emph{during} engagements to maximize total expected discounted reward functions (agent utilities) in an `online' sense (i.e. updating agent behaviors in the middle of an engagement).
While the connections between reinforcement learning and \BO{} have been noted elsewhere \cite{Brochu2010}, the concept and motivations behind using RS and HRMS formulations of \BO{} have not been considered previously. The results provided here show a clear benefit for improving the quality of the surrogate model and reliability of the parameter search process using a minimum number of function evaluations (combat simulations).  
However, while this work showed some analysis of the effects of RS compared to traditional single sampling and multi-point sampling (MS) \BO{} methods, additional theoretical work is needed to fully understand the benefits of RS and HRMS strategies. This deeper understanding can be used to develop techniques for automatically calculating RS and MS sample size portions in HRMS, to ensure best possible optimization performance without guesswork or brute force evaluations to determine the best configurations. The existing literature regarding statistical experiment design may be helpful to understanding the contributions of RS in this regard. 

While this work only focused on simulated engagements between two AIs, the \BO{} methods developed here can in principle be extended to simulations with human participants. This work suggests many interesting pathways for exploring how \BO{} techniques can be used to generate adaptive red agents for human trainees. In addition, the \BO{} methods developed here could be used to develop `backseat coaches' for human blue force operators, i.e. AI decision aids that suggest behaviors that will most likely yield successful results based on observed performance. 
The problem of simultaneously training multiple red and blue force agents presents many other interesting possibilities for extending the \BO{} techniques developed here. To cope with the high dimensionality and sensitivity to hyperparameter settings in such applications, it may be necessary to combine RS and HRMS approaches with other sophisticated GP regression and acquisition function optimization strategies not considered here, such as fully Bayesian hyperparameter learning \cite{Snoek-NIPS-2012} and `random embedding' \BO{} for very high-dimensional problems \cite{Wang2013}.

    \newpage
    \bibliography{References}
\end{document}